\documentclass[sigconf]{acmart}

\setcopyright{none}
\usepackage{todonotes}
\usepackage{booktabs}

\usepackage{url,bm}
\usepackage{latexsym}

\usepackage{bbm}
\usepackage{algorithm}
\usepackage[noend]{algorithmic}
\usepackage{multirow}

\usepackage{graphicx}
\usepackage{subcaption}
\usepackage{dblfloatfix}
\usepackage{url}
\usepackage{breakurl}
\usepackage{hyperref}

\let\svthefootnote\thefootnote

\usepackage{etoolbox}
\newtoggle{conf}
\toggletrue{conf}

\newcommand{\cut}[1]{}

\newcommand\blfootnote[1]{%
  \begingroup
  \renewcommand\thefootnote{}\footnote{#1}%
  \addtocounter{footnote}{-1}%
  \endgroup
}

\renewcommand\footnotetextcopyrightpermission[1]{}
\begin{document}
\title{Fairness-Aware Online Personalization}

\author{G Roshan Lal}
\affiliation{%
  \institution{LinkedIn Corporation}
}
\email{rlal@linkedin.com}

\author{Sahin Cem Geyik$~^{*}$}
\affiliation{%
  \institution{Facebook}
}
\email{scgeyik@fb.com}

\author{Krishnaram Kenthapadi$~^{*}$}
\affiliation{%
  \institution{Amazon AWS AI}
}
\email{kenthk@amazon.com}

\begin{abstract}
Decision making in crucial applications such as lending, hiring, and college admissions has witnessed increasing use of algorithmic models and techniques as a result of confluence of factors such as ubiquitous connectivity, ability to collect, aggregate, and process large amounts of fine-grained data using cloud computing, and ease of access to applying sophisticated machine learning models. Quite often, such applications are powered by search and recommendation systems, which in turn make use of personalized ranking algorithms. At the same time, there is increasing awareness about the ethical and legal challenges posed by the use of such data-driven systems. Researchers and practitioners from different disciplines have recently highlighted the potential for such systems to discriminate against certain population groups, due to biases in the datasets utilized for learning their underlying recommendation models.

We present a study of fairness in online personalization settings involving ranking of individuals. Starting from a \emph{fair} warm-start machine learned model, we first demonstrate that online personalization can cause the model to learn to act in an unfair manner, if the user is biased in his/her responses. For this purpose, we construct a stylized model for generating training data with potentially biased features as well as potentially biased labels, and quantify the extent of bias that is learned by the model when the user responds in a biased manner as in many real-world scenarios. We then formulate the problem of learning personalized models under fairness constraints, and present a regularization based approach for mitigating biases in machine learning. We demonstrate the efficacy of our approach through extensive simulations with different parameter settings.

Our implementation of data model, online learning, fair regularization and relevant simulations can be found here:\\ \href{https://github.com/groshanlal/Fairness-Aware-Online-Personalization}{\emph{https://github.com/groshanlal/Fairness-Aware-Online-Personalization}}
\end{abstract}

\iffalse

\begin{CCSXML}
<ccs2012>
 <concept>
  <concept_id>10010520.10010553.10010562</concept_id>
  <concept_desc>Computer systems organization~Embedded systems</concept_desc>
  <concept_significance>500</concept_significance>
 </concept>
 <concept>
  <concept_id>10010520.10010575.10010755</concept_id>
  <concept_desc>Computer systems organization~Redundancy</concept_desc>
  <concept_significance>300</concept_significance>
 </concept>
 <concept>
  <concept_id>10010520.10010553.10010554</concept_id>
  <concept_desc>Computer systems organization~Robotics</concept_desc>
  <concept_significance>100</concept_significance>
 </concept>
 <concept>
  <concept_id>10003033.10003083.10003095</concept_id>
  <concept_desc>Networks~Network reliability</concept_desc>
  <concept_significance>100</concept_significance>
 </concept>
</ccs2012>  
\end{CCSXML}

\ccsdesc[500]{Computer systems organization~Embedded systems}
\ccsdesc[300]{Computer systems organization~Redundancy}
\ccsdesc{Computer systems organization~Robotics}
\ccsdesc[100]{Networks~Network reliability}

\keywords{ACM proceedings, \LaTeX, text tagging}
\fi

\maketitle
\blfootnote{
\emph{RecSys 2020, FAccTRec Workshop: Responsible Recommendation, Sept 2020},\\
2020. ACM ISBN 978-x-xxxx-xxxx-x/YY/MM. . . \$15.00\\
\href{https://doi.org/10.1145/nnnnnnn.nnnnnnn}{https://doi.org/10.1145/nn\%nnnnn.nnnnnnn}
}

\section{Introduction}\label{sec:intro}

\let\thefootnote\relax\footnotetext{$^{*}$~Work done while the authors were at LinkedIn.}
\let\thefootnote\svthefootnote

Algorithmic models and techniques are being increasingly used as part of decision making in crucial applications such as lending, hiring, and college admissions due to factors such as ubiquitous connectivity, ability to collect, aggregate, and process large amounts of fine-grained data using cloud computing, and ease of access to applying sophisticated machine learning models. Quite often, such applications are powered by search and recommendation systems. In such applications, users may not always be able to articulate their information need in the form of a well-defined query, although they can usually determine if a displayed result (such as a potential candidate for hiring or college admission) is relevant to their information need or not. Hence, it may be desirable to personalize the results to each user based on their (explicit or implicit) responses, ideally in an online fashion. At the same time, human responses and decisions can quite often be influenced by conscious or unconscious biases, in addition to the relevancy of the results, causing the personalized models themselves to become biased. In fact, there is increasing awareness about the ethical and legal challenges posed by the use of such data-driven systems. Researchers and practitioners from different disciplines have highlighted the potential for such systems to discriminate against certain population groups, due to biases in the datasets utilized for learning their underlying recommendation models. Several studies have demonstrated that recommendations and predictions generated by a biased machine learning model can result in systematic discrimination and reduced visibility for already disadvantaged groups \cite{nips_2017_tutorial, dwork_2012, hajian_2016_tutorial, pedreschi_2009}. A key reason is that machine learned models that are trained on data affected by societal biases may learn to act in accordance with them.

Motivated by the need for understanding and addressing algorithmic bias in personalized ranking mechanisms, we perform a study of fairness or lack thereof in online personalization settings involving ranking of individuals. Starting from a {\em fair} warm-start machine learned model\footnote{~ We note that although such an unbiased model may not be available in practice, we show that the personalized model can become highly biased in spite of starting from an unbiased model.}, we first demonstrate that online personalization can cause the model to learn to act in an unfair manner, if the user is biased in his/her responses. For this purpose, we construct a stylized mathematical model for generating training data with potentially biased features as well as potentially biased labels. We quantify the extent of bias that is learned by the model when the user sometimes responds in a biased manner (either intentionally or unintentionally) as in many real-world settings. We study both bias and precision measures under different hyper-parameter choices for the underlying machine learning model. Our framework for generating training data for simulation purposes could be of broader use to evaluate fairness-aware algorithms in other settings as well.
We then formulate the problem of learning personalized models under fairness constraints, and present a regularization based approach for mitigating biases in machine learning, specifically for linear models. We demonstrate the efficacy of our approach through extensive simulations with different parameter settings. To summarize, the contributions of our work are as follows:
\begin{itemize}
\item A stylized mathematical model for simulating the biased behavior of users and generating data that is amenable to learning a biased model, if labeled by such biased users.
\item Demonstration of potential bias in online personalization systems for ranking/recommending individuals.
\item A regularization based strategy to mitigate algorithmic bias during model training. 
\end{itemize}
The rest of the paper is organized as follows. We first present a mathematical model for studying bias in online personalization systems and the associated experimental results (\S\ref{sec:studypersonalizationbias}). Next, we propose a regularization based method to mitigate algorithmic bias and demonstrate its effectiveness through an empirical study (\S\ref{sec:fair_regularization}). Finally, we present the related work in \S\ref{sec:relatedwork} and conclude the paper in \S\ref{sec:conclusion}.

\section{Related Work} \label{sec:relatedwork}
Algorithmic bias, discrimination, and related topics have been studied across disciplines such as law, policy, and computer science \cite{nips_2017_tutorial, hajian_2016_tutorial}. Two different notions of fairness have been explored in many recent studies: (1) {\em individual fairness}, which requires that similar people be treated similarly~\cite{dwork_2012}, and (2) {\em group fairness}, which requires that the disadvantaged group be treated similarly to the advantaged group or the entire population~\cite{pedreschi_2008}. There is extensive work on identifying and measuring the extent of discrimination (e.g.,~\cite{angwin_2016, caliskan_2017, pedreschi_2008}), on mitigation approaches in the form of fairness-aware algorithms (e.g.,~\cite{calders_2010, celis_2017, corbett_2017, dwork_2012, friedler_2016, friedler2018comparative, hajian_2013, hardt_2016, jabbari_2017, kamishima_2012, kleinberg_2017, woodworth2017learning, zafar2017fairness, zehlike_2017, zemel_2013}), and on inherent trade-offs in achieving different notions of fairness~\cite{corbett_2017, dwork_2012, friedler_2016, kleinberg_2017}.

~

\noindent {\it Modifying Model Training Algorithm for Fairness}:
While several studies have focused on data transformation/representation (pre-processing, post-processing, or during training) for ensuring fairness (e.g.,~\cite{feldman_2015, zemel_2013, edwards_2016, louizos_2016}), there has also been work on modifying the model training algorithm itself. Regularization based methods for fairness have been proposed in~\cite{kamishima_2012, suay_2017,mary2019}. To meet the independence constraint of fairness, an approximate estimate of mutual information of the classifier output and the protected attribute is computed and used as a regularizer in~\cite{kamishima_2012}. This approach is extended in \cite{mary2019} for continuous protected attributes. Correlation between the protected attribute and the classifier output is further used as the regularizer in \cite{suay_2017}. While we use a similar correlation based approach for performing classification, our method differs in the following key aspects. We first train an auxiliary model for predicting the protected attribute in terms of other features, and then guide our main model to be uncorrelated with this auxiliary model. Our problem setting is also slightly different. We use online learning wherein the training samples are not picked uniformly at random, but in a special way where we only show the highest scored candidate to the user for feedback in each iteration. The idea of auxiliary model has earlier been used in the context of training an adversarial neural network in a fair manner~\cite{beutel_2017}, but our training method is significantly different. While the auxiliary model is trained to underperform in~\cite{beutel_2017}, we train the auxiliary model to perform well and use it to steer the main model away from it. Another recent paper \cite{romanov2019} penalizes the covariance between the probability assignment of a user to its correct label, and his/her name's embedding, without access to protected attributes.

Some recent work \cite{beutel2019,narasimhan2020} have also dealt with achieving fairness via regularization utilizing pair-wise comparisons where paired items are chosen from different protected attribute groups. Both of these works in general attempt learn models which have similar probability to assign a larger score to better items than a worse item, regardless of the protected attribute.

Finally, a meta-learning approach through regularization is proposed in \cite{slack2020}, where the authors propose to learn a model that performs well and fair across a set of tasks, and use a small size sample for each task. Within each task they apply a regularization term towards getting the fair model which is based on the average probability that a class is to be classified as positive given the protected attribute (parity), or given the protected attribute and label (equalized odds or equal opportunity).

{\it Fairness in Ranking}:
There has been a lot of recent work on fairness in ranking~\cite{celis_2017, yang_2017, zehlike_2017, singh_2018, biega_2018, asudeh_2017, morik_2020}. Algorithms for reranking subject to fairness constraints, specified as the maximum (or minimum) number of elements of each class that can appear at any position in the ranking, have been proposed in~\cite{celis_2017, zehlike_2017}. Fairness measures for ranking have been proposed in~\cite{yang_2017}. Fairness constraints are modeled as linear constraints along with a optimizing for relevance in ~\cite{celis_2017}. Fairness in ranking and feedback from it are modeled as a stochastic bandit problem in ~\cite{singh_2018}. Achieving fairness in a dynamic learning to rank setting using a fair controller is discussed in ~\cite{morik_2020}.

Refer Appendix \ref{sec:background} for a discussion on legal framework on discrimination and relevant notions of fairness popular in machine learning.

\section{Study of Fairness in Online Personalization}\label{sec:studypersonalizationbias}

We next present a study of fairness in online personalization settings involving ranking of individuals. Starting from a {\em fair} warm-start machine learned model, we first demonstrate that online personalization can cause the model to learn to act in an unfair manner, if the user is biased in his/her responses. For this purpose, we construct a stylized model for generating training data with potentially biased features as well as potentially biased labels, and quantify the extent of bias that is learned by the model when the user responds in a biased manner as in many real-world scenarios (\S\ref{sec:data}). We then present experimental results from our simulation framework, where we consider both bias and precision measures under different hyper-parameter choices for the underlying machine learning model (\S\ref{sec:experiment1} -- \S\ref{sec:evolution}).
We chose to use a simulation framework since it allows us to model {\em biased users}, and further vary their levels of bias.

\subsection{Data Model} \label{sec:data}
We next present our stylized model for generating training data which captures some of the sources of unfairness that is usually present in real-world data, as listed in \S\ref{sec:biassources}. We split our discussion into two parts. We describe how we generate feature vectors in \S\ref{sec:features} and we describe how we further label the data in \S\ref{sec:labels}.

\subsubsection{Features}\label{sec:features}
\begin{figure}[!t]
	\centering
	\includegraphics[width=3.3in]{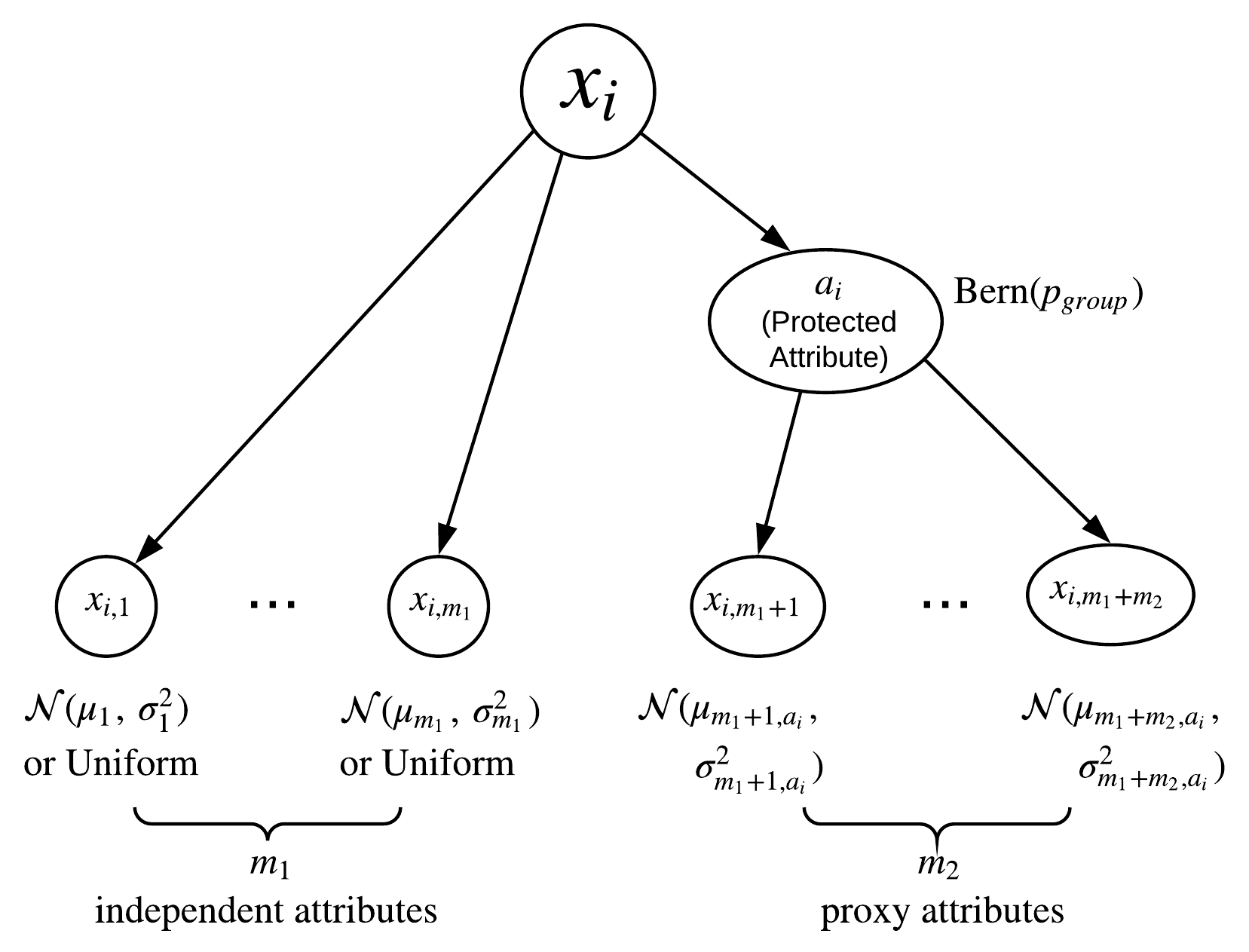}
	\caption{Model to simulate feature vectors of training data.}
	\label{fig:bayesian-network}
\end{figure}

Suppose that our training data consists of $n$ data points, where the $i^{th}$ data point $x_i$ is a vector consisting of $m$ seemingly harmless attributes: $x_i$ = ($x_{i,1}~,~x_{i,2}~,~x_{i,3}~,~ \ldots ~,~x_{i,m}$).
For the $i^{th}$ data point, let $a_i$ denote the corresponding value of the protected attribute. Of the $m$ attribute values of $x_i$, let the first $m_1$ attribute values be independent of the protected attribute value $a_i$ and the remaining $m_2$ attribute values be generated in a fashion dependent on $a_i$ ($m = m_1 + m_2$). We can thus model the $m$ attributes of each data point as being generated according to the scheme in Figure~\ref{fig:bayesian-network}. As presented in the figure, for each $i$, we first generate the protected attribute $a_i$ independently using a Bernoulli random variable which gives $1$ with probability $p_{group}$ and $0$ with probability $1 - p_{group}$. We can interpret $a_i$ as encoding the membership to some protected class like belonging to a particular gender, race, or age $\ge$ 40, with $a_i = 1$ as corresponding to the privileged group. For the other (non-protected) attributes, we choose those that are ``harmless'' (independent of the protected attribute, i.e., $x_{i,1}$ through $x_{i,m_1}$) to be drawn either from a uniform distribution between $0$ and $1$, or a normal distribution with their respective means and variances, independently. For the ``proxy attributes'' (dependent on the protected attribute, i.e., $x_{i,m_1+1}$ through $x_{i,m_1+m_2}$), we draw them independently from normal distributions, whose means and variance are functions of the protected attribute value $a_i$. Although we describe the attribute values to be drawn from either a uniform distribution or a normal distribution for simplicity, we remark that our model can easily be extended to allow more general distributions, e.g., categorical distributions for categorical attributes.

\textbf{\emph{Current Setting:}} While the above mathematical framework is quite general, in our experiments for the current work, we chose $m_1 = 1$ and $m_2 = 2$ so that $m = 3$, for simplicity. We therefore employed one harmless attribute which we draw uniformly in [0,1], and two proxy attributes which we draw from Normal distributions dependent on the protected attribute. Finally, we held the variance constant for all Normal distributions, although as discussed above, this does not necessarily have to be so.

\subsubsection{Labels}\label{sec:labels}
\begin{figure}[!t]
\centering
\includegraphics[width=2.8in]{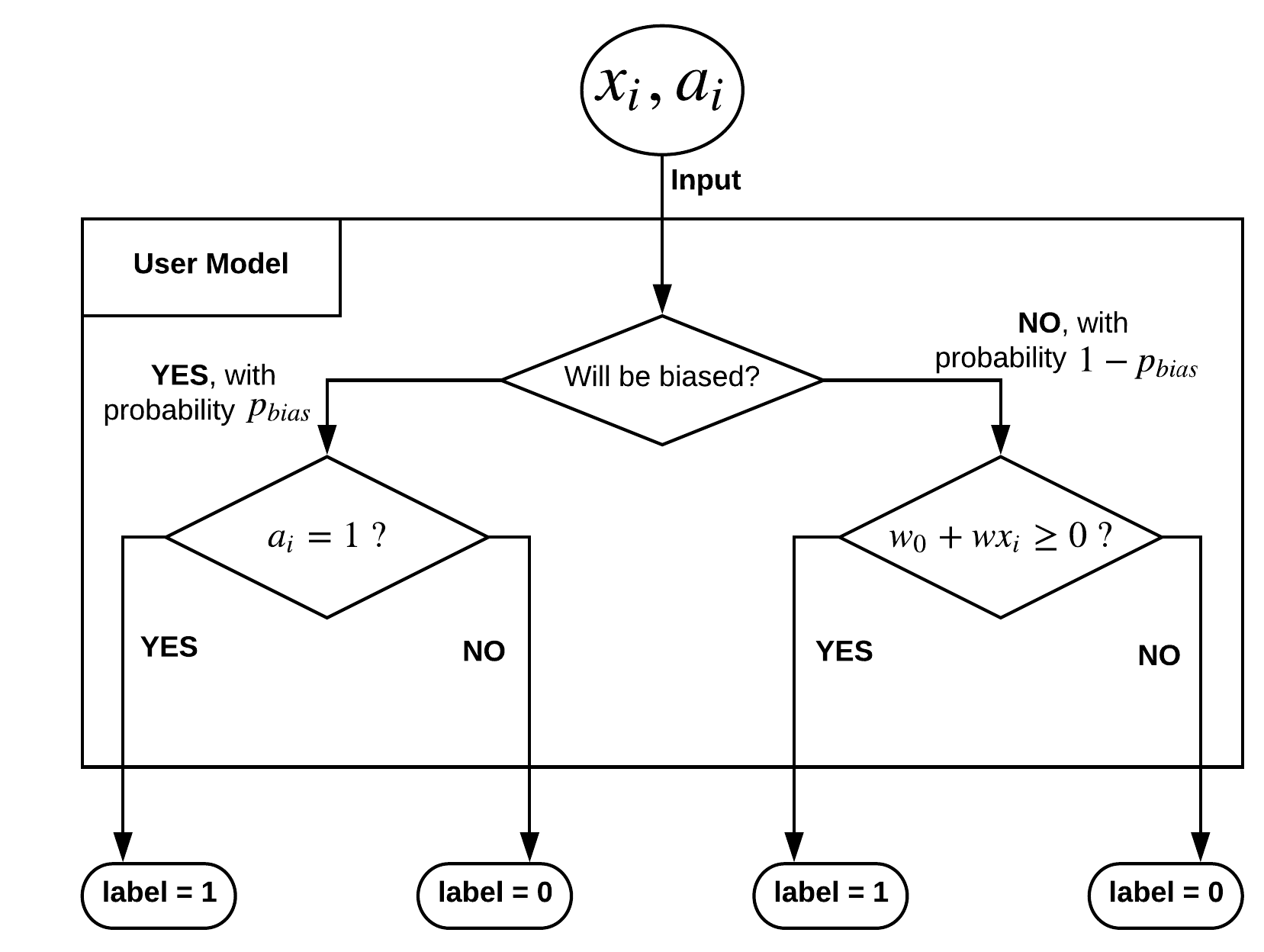}
\caption{Model to label simulated feature vectors.}
\label{fig:user-model}
\end{figure}

As stated earlier, our simulation framework enables us to model the behavior of {\em biased users} and vary their levels of bias, which we cannot obtain from real world datasets.
Our labeling strategy for a potentially biased user is presented in Figure~\ref{fig:user-model}, and is as follows. Given the feature vector associated with a data point $x_i$, we need to label $x_i$ as either ``Accept'' ($1$) or ``Reject'' ($0$). First, we determine whether the user who labels this data point would behave in a biased or unbiased manner, using an independent Bernoulli trial with parameter $p_{bias}$. This means that the user chooses to behave in a biased (unfair) manner with probability $p_{bias}$, and in an unbiased (fair) manner with probability $1-p_{bias}$.

If the user chooses to be fair, he/she takes a fixed linear combination of the attribute values in the feature vector ($w_0 + \sum_{k=1}^{m} w_k x_{i,k}$, or in short, $w_0 + w^T x_i$) and labels the data point according to the sign of the linear combination (1 if the linear combination is non-negative, and 0 if it is negative). If the user chooses to be unfair, he/she labels the data point $x_i$ purely based on the protected attribute value, $a_i$. More precisely, the user labels as ``Reject'' (0) whenever $a_i = 0$ and ``Accept'' (1) whenever $a_i = 1$.

The weights $[w_0, w_1, \ldots, w_m]$ are constants as defined by the user and unknown to the machine learning model which subsequently uses the data. The user chooses these weights without any knowledge of how the data is generated and has no knowledge of which attributes of the feature vector are independent of the protected attribute. Hence, the way the data is generated, the way it is labeled, and the way it is used for training the machine learning model are all blind to each other. 
 
In the above model of simulating feature vectors and labeling them, we have captured the key aspects of unfairness listed in \S\ref{sec:background}:
\begin{itemize}
	\item \emph{Proxy Attributes:} $x_{i,m_1 + 1} ~ , ~ \ldots ~ , ~ x_{i,m_1 + m_2}$ serve as proxy attributes.
	\item \emph{Quantity of training data for different values of the protected attribute:} This is controlled by the parameter $p_{group}$ while generating the feature vectors.
	\item \emph{Quality of training data for different values of the protected attribute:} This pertains to how the proxy attributes are generated from the protected attribute. 
Depending on the functions used for mapping the protected attribute value to the parameters of the distribution used for generating values of the proxy attributes, we may obtain training data with different quality for different groups. For example, the variance of the normal distributions used in our setting could be small for $a_i = 0$ and large for $a_i = 1$.
	\item \emph{Noisy/biased labels in training data:} This is controlled by parameter $p_{bias}$ which decides the user's level of unfairness. When the user behaves in an unfair manner, the resulting labels are biased, with $a_i = 0$ resulting in ``Reject'' (0) and $a_i = 1$ resulting in ``Accept'' (1).
\end{itemize}
We remark that our stylized training data generation model could be extended/generalized to incorporate a combination of ``objectivity'' and (possibly unconscious) biases, which may be closer to be more likely in practice. For instance, instead of considering the user to be either fully objective or fully biased when labeling, we could model the user as say, boosting the ``objective'' score based on the protected attribute value when determining whether to accept or reject. Another extension to our model would be to allow for the user to be partially biased: when the user is biased, we can view the label assigned by the user as a random variable dependent on the protected attribute value, and draw the value of the label from a Bernoulli distribution whose parameter could be influenced by the protected attribute value. For simplicity, we limit ourselves to the stylized model in this paper.

\subsection{Experimental Setup}\label{sec:experiment1}
Next, we would like to describe our experimental setup  to measure how different choices of hyper-parameters like learning rate and the number of training rounds affect fairness and precision in the outcomes of the online learning model.

\emph{Online Learning}: In every iteration, we score all the data points in the training data and obtain the next data point for training by choosing the data point with the maximum score (highest possibility of being labelled as ``Accept'' as per the current model) that has not been used for training till now. The underlying setting is that we present one data point at a time to the user (we select the data point with the highest score, amongst those that have not been shown to the user so far), to which the user responds either positively or negatively. In either case, the online personalization model attempts to incorporate such user preference, so that a more personalized result can be presented next time to the user. 
We use the feature vectors and label of this new data point for training in the next round and this is repeated over and over again. The algorithm is started using a ``warm-start'' model in which data from a fair source is used to train the models by randomly sampling training data points from the fair dataset a number of times. We assume the availability of such a fair dataset so that we can then compare our final biased model against the warm-start baseline. 

\emph{Our Setup}:
First, we generate 12000 data samples\footnote{~~ As discussed in \S\ref{sec:features}, for our experiments, we utilize one ``harmless'' attribute ($x_{i,1}$ for each $x_i$), and two ``proxy'' attributes ($x_{i,2}$ and $x_{i,3}$ for each $x_i$). In our training data generation, for each data point, we choose the second and the third features to be dependent on the protected attribute value, but in opposite directions. For each $i$, if $a_i = 0$, we draw $x_{i,2}$ independently from the normal distribution with mean 0.35 and standard deviation 0.12, and if $a_i = 1$, we draw $x_{i,2}$ independently from the normal distribution with mean 0.65 and standard deviation 0.12. The mean values are reversed for the third attribute $x_{i,3}$. $x_{i,1}$ is drawn from the uniform distribution over [0,1].} at $p_{group} = 0.5$ and label them\footnote{~~ We utilized the following weights for the user label decision model (\S\ref{sec:labels}): $w_0 = -0.48$ (intercept), $w_1 = 0.35$, and $w_2 = w_3 = 0.28$. For both feature and label generation, we experimented with different choices of the parameters and observed qualitatively similar results.} in a fair manner (that is, $p_{bias}$ is chosen to be $0$ in \S\ref{sec:labels}, so that the labeling is based on the sign of the linear combination). Our model is a online linear model (perceptron) with the update equation:
$$w_{n+1} = w_{n} + \eta(y_n - \text{sgn}(w_n^Tx_n))x_n$$
where $\eta$ is the learning rate.
We randomly sample 1000 data points from this set and use these data points to train a perceptron. We call this perceptron the \textbf{``warm-perceptron''} from here on. The data points that were generated are discarded beyond this point.

\begin{figure*}[!t]
    \centering
    \begin{subfigure}[b]{1.6in}
        \includegraphics[width=1.6in]{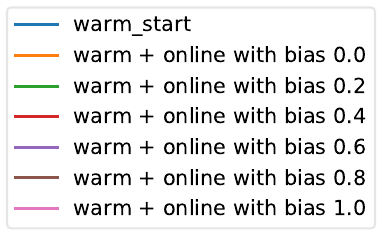}
        \caption{Legend}
        \label{fig:skew_legend}
    \end{subfigure}
    \quad \begin{subfigure}[b]{2.2in}
        \includegraphics[width=2.2in]{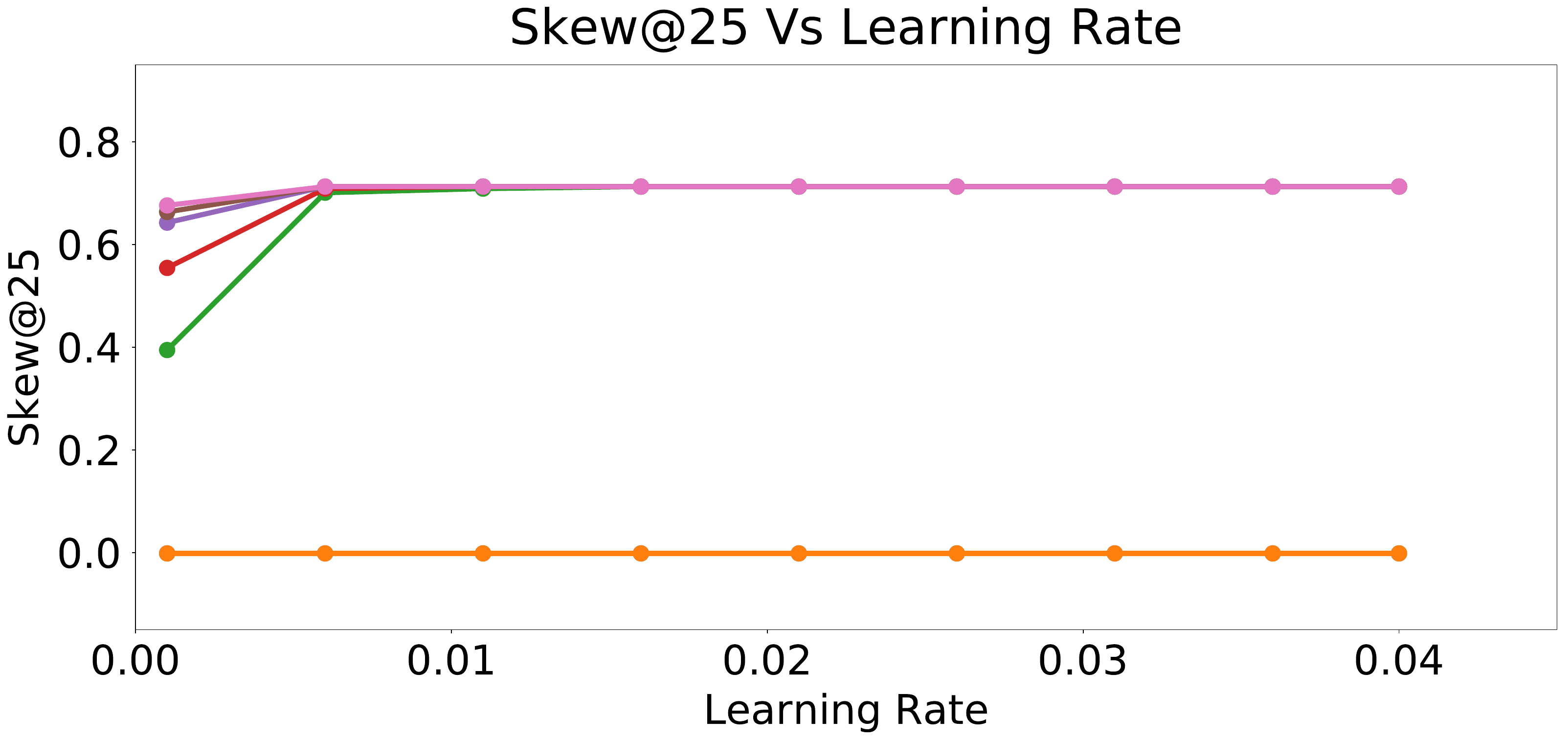}
        \caption{Skew$_{group = 1}$@25}
        \label{fig:skew25}
    \end{subfigure}
    \quad \begin{subfigure}[b]{2.2in}
        \includegraphics[width=2.2in]{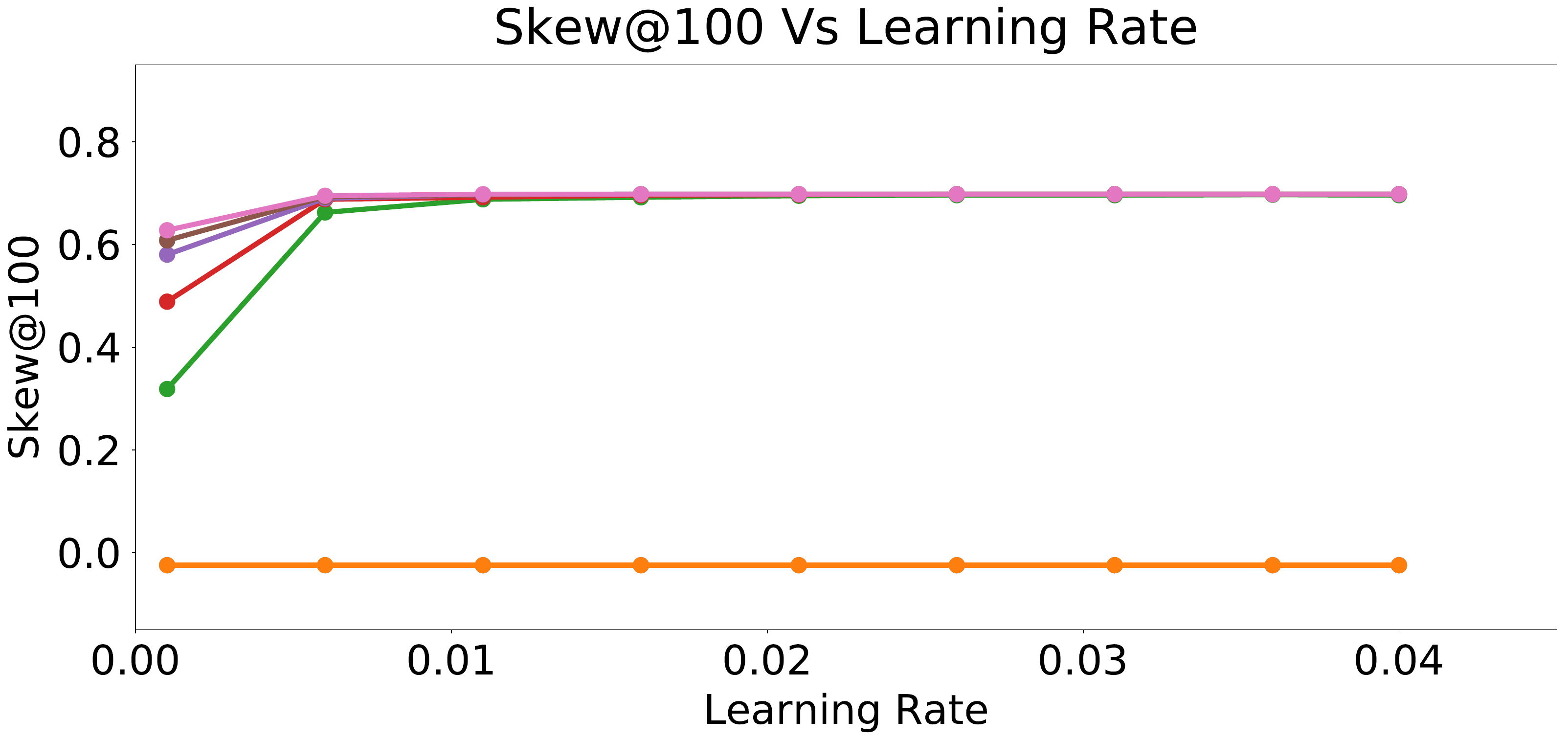}
        \caption{Skew$_{group = 1}$@100}
        \label{fig:skew100}
    \end{subfigure}
    
    \hfill
    
    \begin{subfigure}[b]{2.2in}
        \includegraphics[width=2.2in]{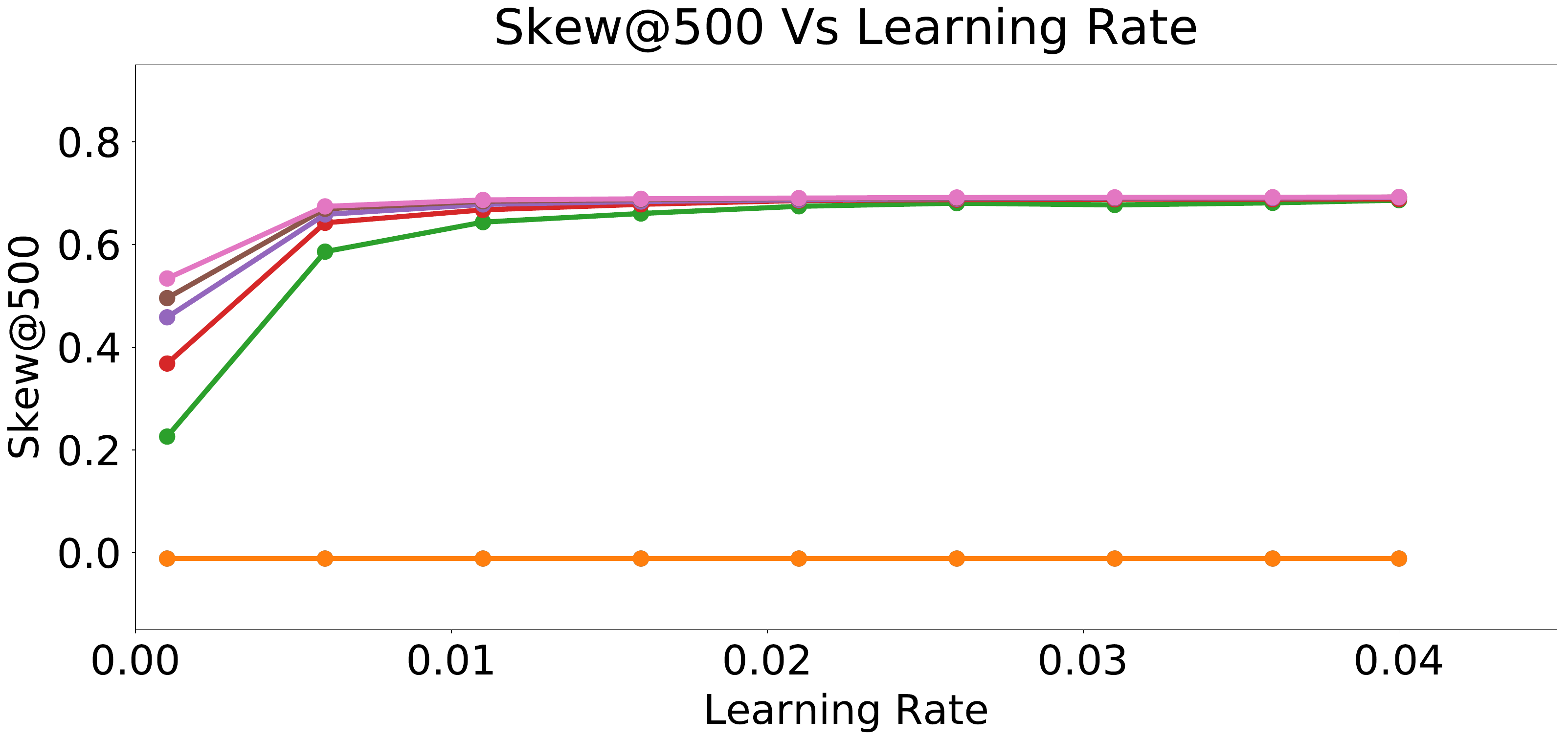}
        \caption{Skew$_{group = 1}$@500}
        \label{fig:skew500}
    \end{subfigure}
    \quad \begin{subfigure}[b]{2.2in}
        \includegraphics[width=2.2in]{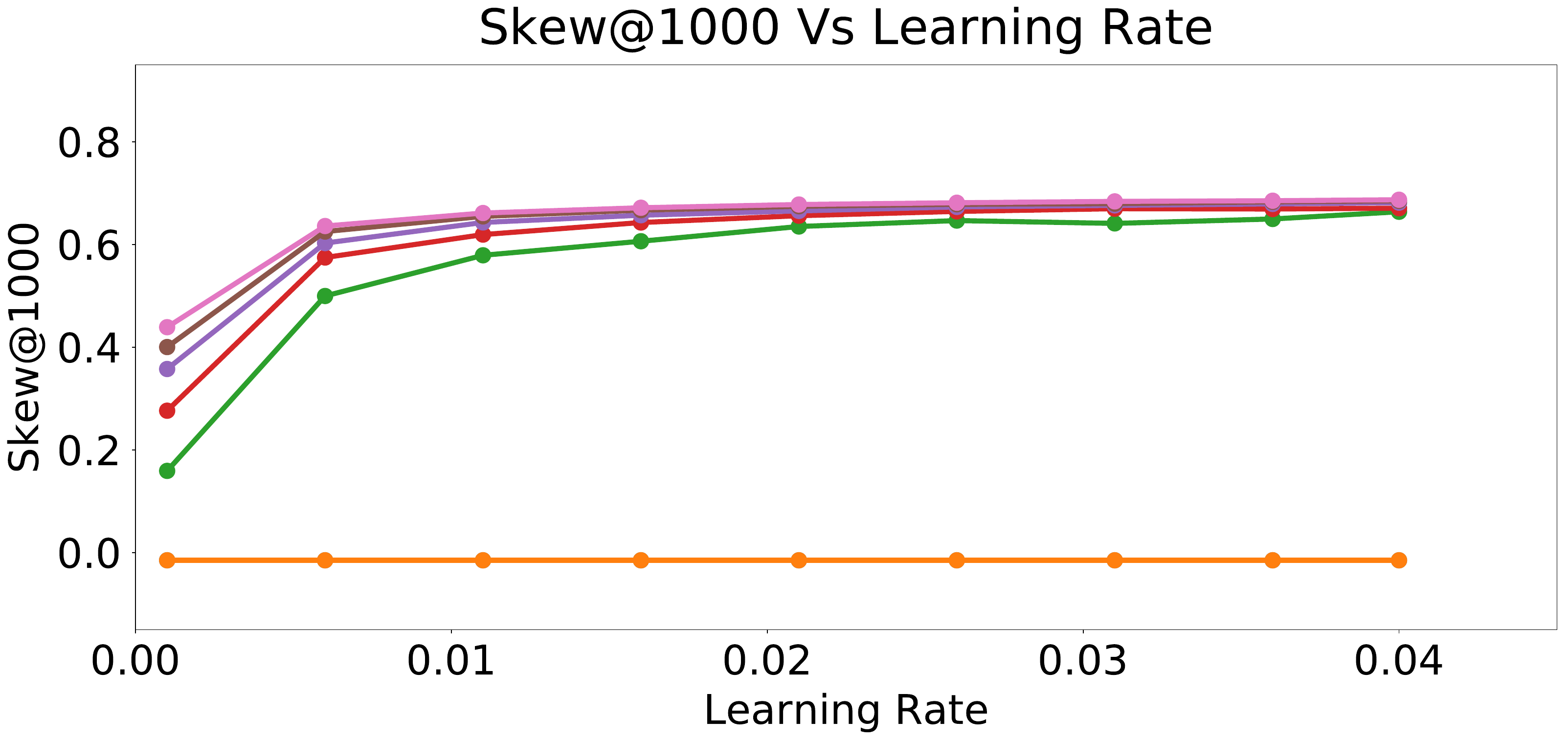}
        \caption{Skew$_{group = 1}$@1000}
        \label{fig:skew1000}
    \end{subfigure}
    \quad \begin{subfigure}[b]{2.2in}
        \includegraphics[width=2.2in]{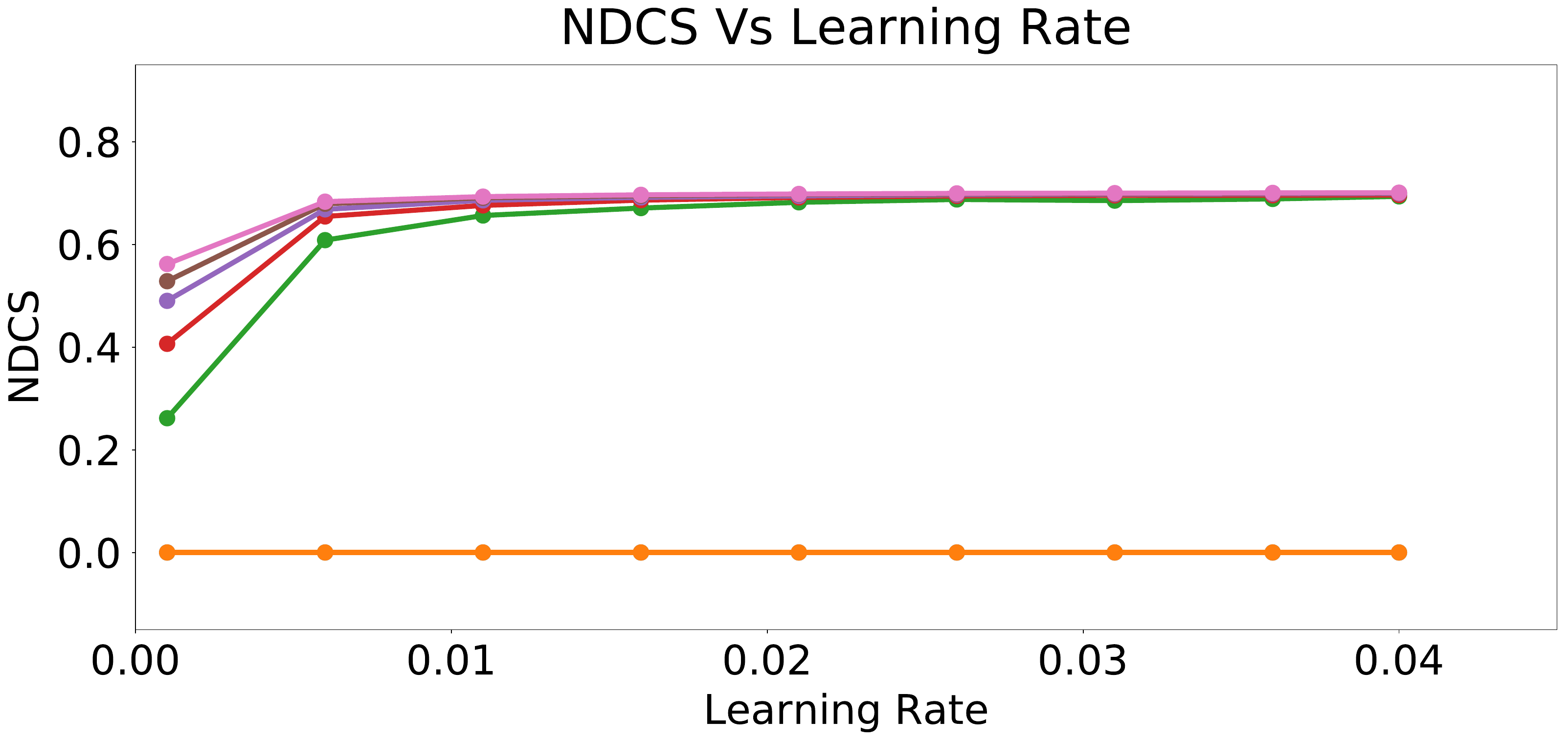}
        \caption{NDCS}
        \label{fig:ndcs}
    \end{subfigure}
    
    \caption{Skew and NDCS Results for the Final Model After Online Personalization Updates}\label{fig:skew}
\end{figure*}

Next, we generate 12000 data samples at $p_{group} = 0.5$ and label them with the bias parameter set to a specific choice of $p_{bias}$. We start from the warm-perceptron model. In each round, we score all the 12000 data points using perceptron weights. We pick the point with highest score (the data point that is farthest from the perceptron on the positive side of the half-plane) that is not yet used for training, obtain its label (following \S\ref{sec:labels}), and then update the warm-perceptron with this data point and its label.

\emph{Metrics}: Our measures for evaluating bias are based on the assumption that the distribution of the protected attribute values for the top ranked candidates (i.e., based on the order with which a candidate gets to be shown to the user in our online personalization setting) should ideally reflect the corresponding distribution over the set of high quality or relevant candidates.

Our first measure, Skew$_{v}$@k computes the logarithmic ratio of the proportion of candidates having $v$ as the value of the protected attribute among the set of $k$ highest scoring candidates to the corresponding proportion among the set of relevant candidates:
\begin{equation}
\label{eq:skew_metric}
\textrm{Skew}_v\textrm{@k}= \log_e\left(\frac{p_{v@k}}{p_{v,\textrm{qualified}}}\right) =\log_e\left(\frac{~~~~~\frac{\sum_j \mathbbm{1}(a_j = v ~ \& ~ \textrm{index}(j) < k)}{k}~~~~~}{~~~~~\frac{\sum_j \mathbbm{1}(a_j = v ~ \& ~ w_0 + w^T x_j \geq 0)}{\sum_j \mathbbm{1}{(w_0 + w^T x_j \geq 0)}}~~~~~}\right) ~,
\end{equation}
where index($j$) denotes the position of candidate $j$ when ordered by decreasing scores, and $p_{v@k}$ and $p_{v,\textrm{qualified}}$ denote the fraction of candidates with protected attribute value $v$ among the top-k recommended candidates and the overall qualified pool of candidates, respectively. Due to our simulation framework, we have further defined being ``qualified'' as being able to pass the linear criteria of the ``fair'' user ($w_0 + w^T x_j \geq 0$). This is one key advantage of utilizing the proposed mathematical framework for our experiments, since understanding which candidate is qualified is otherwise an extremely difficult task.

For each possible value $v$ of the protected attribute, a negative skew corresponds to lesser representation of candidates with value $v$, while a positive skew corresponds to higher representation; a skew of zero is ideal.

Our second measure, $NDCS_{v}$ (Normalized Discounted Cumulative Skew) is a ranking version of the skew measure, inspired by \emph{Normalized Discounted Cumulative Gain} \cite{jarvelin_2002} and the related bias quantification measures proposed in \cite{yang_2017}. We define $NDCS_{v}$ as the cumulative skew over all values of $k$ up to the total number of candidates shown, where each $Skew_{v}@k$ is weighted by a logarithmic discount factor based on $k$, and the sum is normalized by dividing by the sum of the discounting factors:
\begin{equation}
\label{eq:ndcs_metric}
\textrm{NDCS}_v=\frac{1}{Z} ~ \sum_{j}\frac{1}{\log_2(j + 1)} \textrm{Skew}_v\textrm{@j} ~ ,
\end{equation}
where,
\[
Z = \sum_{j} \frac{1}{\log_2(j + 1)} ~ .
\]
This measure is based on the intuition that achieving fairness in top results is more important than at the bottom, since the user is more likely to see and ``Accept'' (or even just be shown) the candidates at the top positions.

\subsection{Evaluation of the Final Model After Online Personalization Updates}\label{sec:fairness}
We first present the results of our study after a round of online personalization update scheme is applied. As discussed in \S\ref{sec:experiment1} (see \emph{Our Setup}), the initial model is what we call a warm-perceptron model, and we apply the online updates for 1000 candidates. At each new candidate, we choose the best candidate for the current model, get the feedback from the user (\S\ref{sec:labels}) and update the current model. At the end of 1000 rounds (candidates, feedback, and updates), we have generated a personalized model. In this section, we compare this personalized model (achieved by the end of $1000^{th}$ update) to the warm-perceptron model (which was the initial model before online updates), via an independent ranking of a large pool of candidates generated per \S\ref{sec:data}, and checking the effect on the fairness and precision metrics.

\begin{figure*}[!t]
    \centering
    \begin{subfigure}[b]{1.6in}
        \includegraphics[width=1.6in]{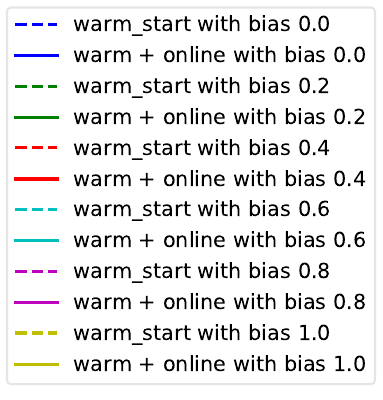}
        \caption{Legend}
        \label{fig:precision_legend}
    \end{subfigure}
    \quad \quad \begin{subfigure}[b]{5.0in}
        \begin{subfigure}[b]{2.3in}
            \includegraphics[width=2.3in]{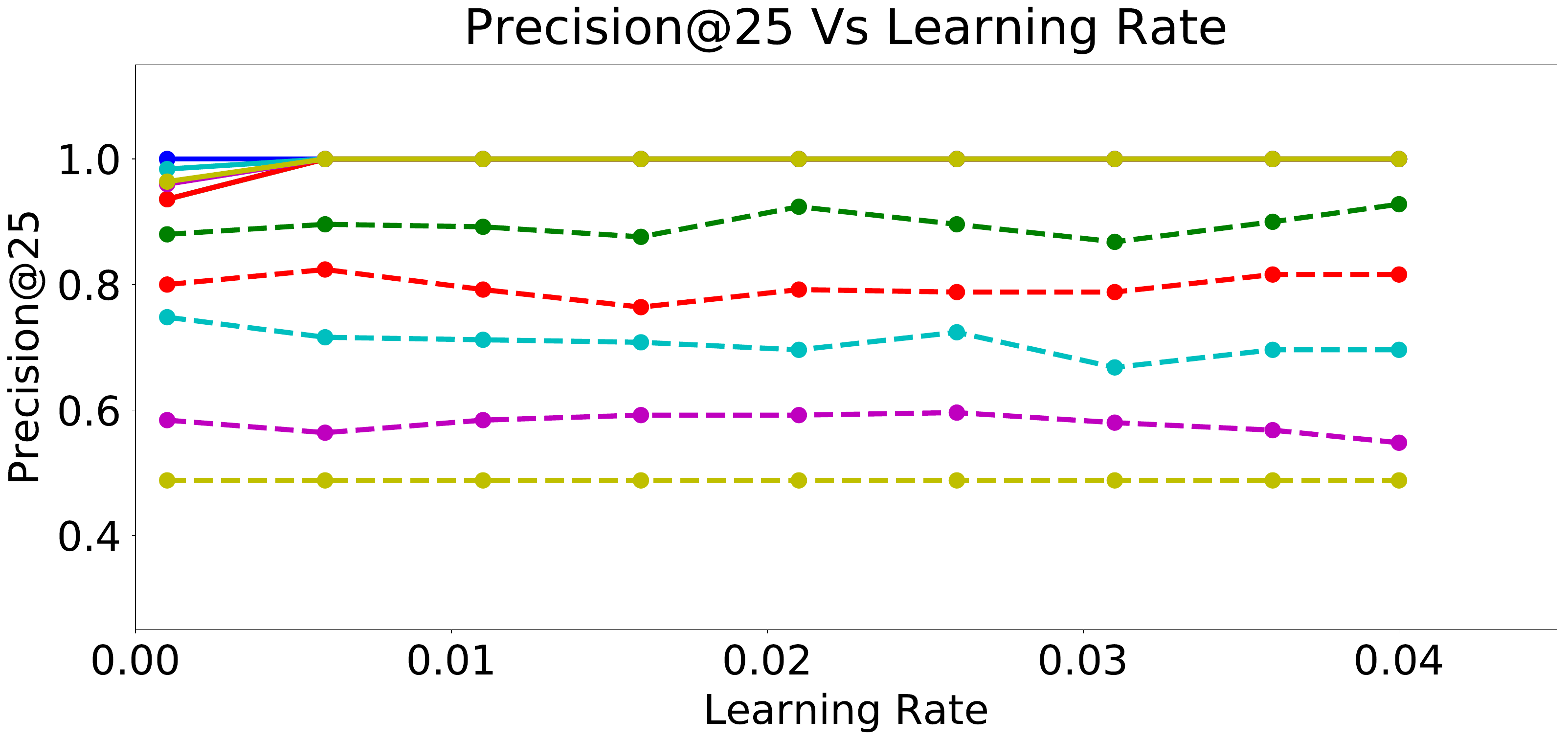}
            \caption{Precision\cut{$_{group = 1}$}@25}
            \label{fig:precision25}
        \end{subfigure}
        \quad \begin{subfigure}[b]{2.3in}
            \includegraphics[width=2.3in]{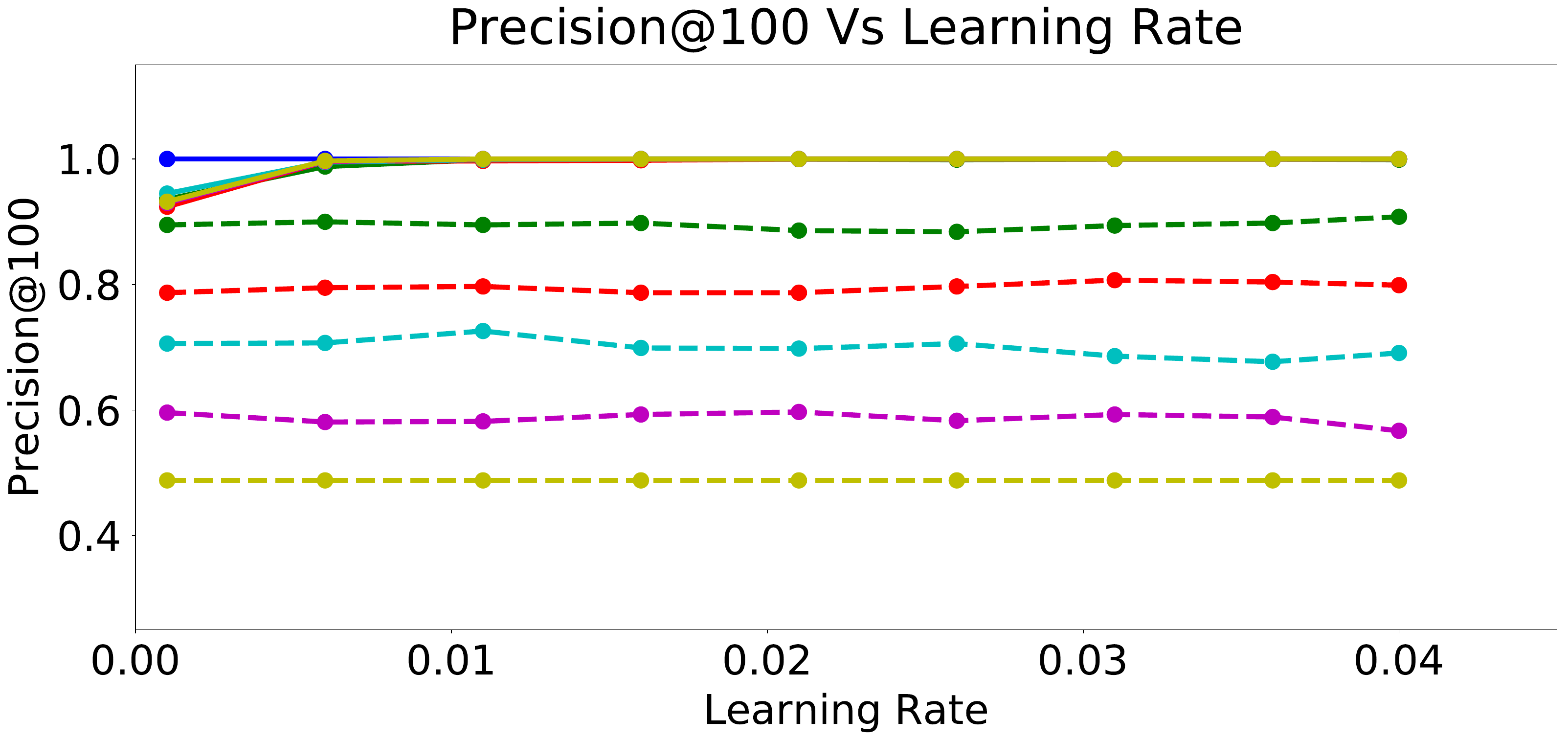}
            \caption{Precision\cut{$_{group = 1}$}@100}
            \label{fig:precision100}
        \end{subfigure}
        
        \hfill
        
        \begin{subfigure}[b]{2.3in}
            \includegraphics[width=2.3in]{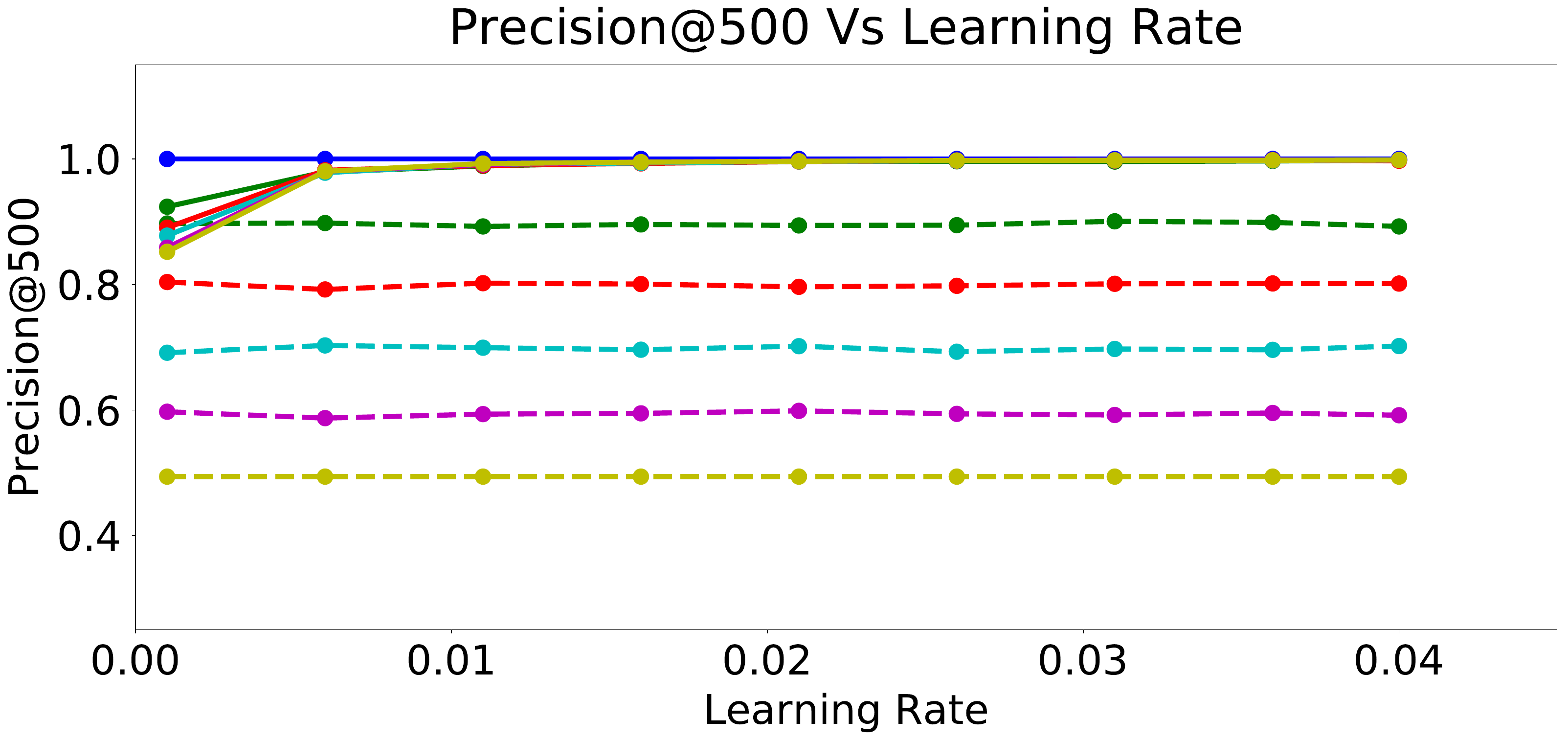}
            \caption{Precision\cut{$_{group = 1}$}@500}
            \label{fig:precision500}
        \end{subfigure}
        \quad \begin{subfigure}[b]{2.3in}
            \includegraphics[width=2.3in]{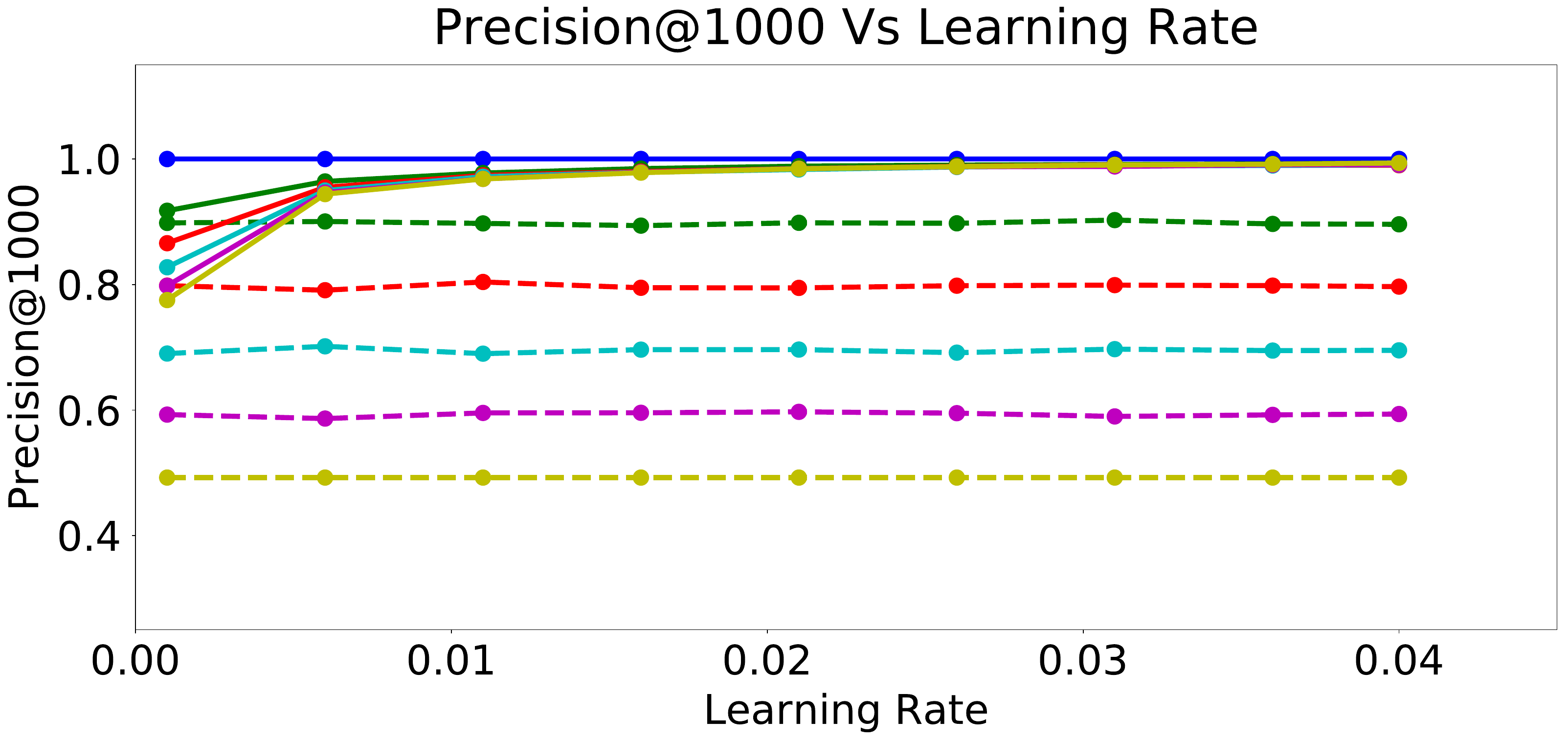}
            \caption{Precision\cut{$_{group = 1}$}@1000}
            \label{fig:precision1000}
    \end{subfigure}
    \end{subfigure}
    \caption{Precision Results for the Final Model After Online Personalization Updates}\label{fig:precision}
\end{figure*}

Since the calculation of the fairness measures like $Skew@k$ (Eq.~\ref{eq:skew_metric}) and $NDCS$ (Eq.~\ref{eq:ndcs_metric}) requires the knowledge of the protected attribute distribution in the baseline population, we use the same data points used for online training and label them using a user model with $p_{bias} = 0$ and the same weights $[w_0, w_1, \ldots, w_m]$ for the purpose of calculating the baseline proportions (similar to the computation in Eq.~\ref{eq:skew_metric}). We use these labels from the fair user to get the baseline proportions (the binary protected attribute proportions in the qualified candidate set) of the two groups.

In Figure \ref{fig:skew}, we present Skew@k and NDCS results for the online personalization updates if we received the feedback to the recommended candidates from users with different $p_{bias}$ values. We observe that as learning rate increases\cut{ from 0 to 0.1}, the skew value increases at first and quickly saturates with increasing learning rate, for all types of users. From Figures \ref{fig:skew25}-\ref{fig:ndcs}, we can see this pattern across all k for Skew@k and also in the NDCS metric. Furthermore, we see that the fairness metrics in \emph{warm\_start} (warm-perceptron) and \emph{warm + online} (online personalization updates applied to warm-perceptron) with 0 bias are exactly the same. The reason is that the warm-perceptron model is also trained with data having 0 bias ($p_{bias}=0$). Thus, the perceptron weights don't change further during the online rounds, since the weights are already learned in the warm-start phase. Hence, warm-start model and warm-start + online with 0 bias correspond to exactly the same perceptron weights. Furthermore, we observe that as the bias in training data increases, the Skew/NDCS metrics also increase. This can be explained as the perceptron learning the unfair behavior of the user by adjusting the weights of the attributes $x_2$ and $x_3$ which encode information about the sensitive attribute.

In Figure \ref{fig:precision}, we examine the precision of the warm-perceptron model vs. the subsequently online trained model on the biased user data at various $p_{bias}$ values, as a function of the learning rate. We observe that as learning rate increases\cut{ from 0 to 0.1}, the precision increases at first and quickly saturates with increasing learning rate. From Figures \ref{fig:precision25}-\ref{fig:precision1000}, we can see this pattern across all k for Precision@k (i.e., precision within the first k candidates recommended for users with different $p_{bias}$). Furthermore, we see that the warm-start model has very low precision while the online trained model has precision close to 1. This is because the warm-start model is trained on data labeled by a user with $p_{bias}=0$ and is not precise enough to predict the response of users with large $p_{bias}$ to the recommended candidates. The online model however learns the unfair behavior of user well enough for top k data points and hence precision@k is significantly improved. This means that online personalization, as expected, comes with a significant advantage for the ``relevance'' of recommendations, especially because it incorporated the algorithmic bias. This demonstrates the potential problem with applying such a preference understanding methodology, even when the initial model learned from a large pool of (say, unbiased) users is fair in recommendations.

\subsection{Study of Personalized Model Evolution with Respect to Fairness and Precision}\label{sec:evolution}
In this experiment, we interrupt the online learning process every 25 rounds and measure the fairness metrics for the ranked list of data points that has been shown thus far. Note that, semantically, this is different from the experiments presented in the previous section where we used the perceptron that is the outcome of the online updates over 1000 rounds (each round is presenting a candidate using the updated model, getting feedback, and further updating the model for the next round), and reranked all the 12000 data points to measure Skew@k, NDCS, and Prec@k. However, in this experiment, we interrupt the online learning process every 25 rounds and use the recommendations so far to measure the metrics. This can be seen as investigating the algorithmic bias that is being integrated into the online personalized model in real-time. Thus, this experiment measures the evolution of fairness in recommendations as we update the model with immediate feedback.

\begin{figure}[!t]
    \centering
    \begin{subfigure}[b]{0.85in}
        \includegraphics[width=0.85in]{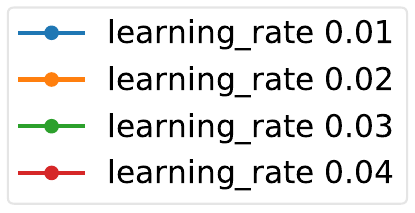}
        \caption{Legend (fixed bias)}
        \label{fig:evolution2}
    \end{subfigure}
    \quad \begin{subfigure}[b]{2.3in}
        \includegraphics[width=2.3in]{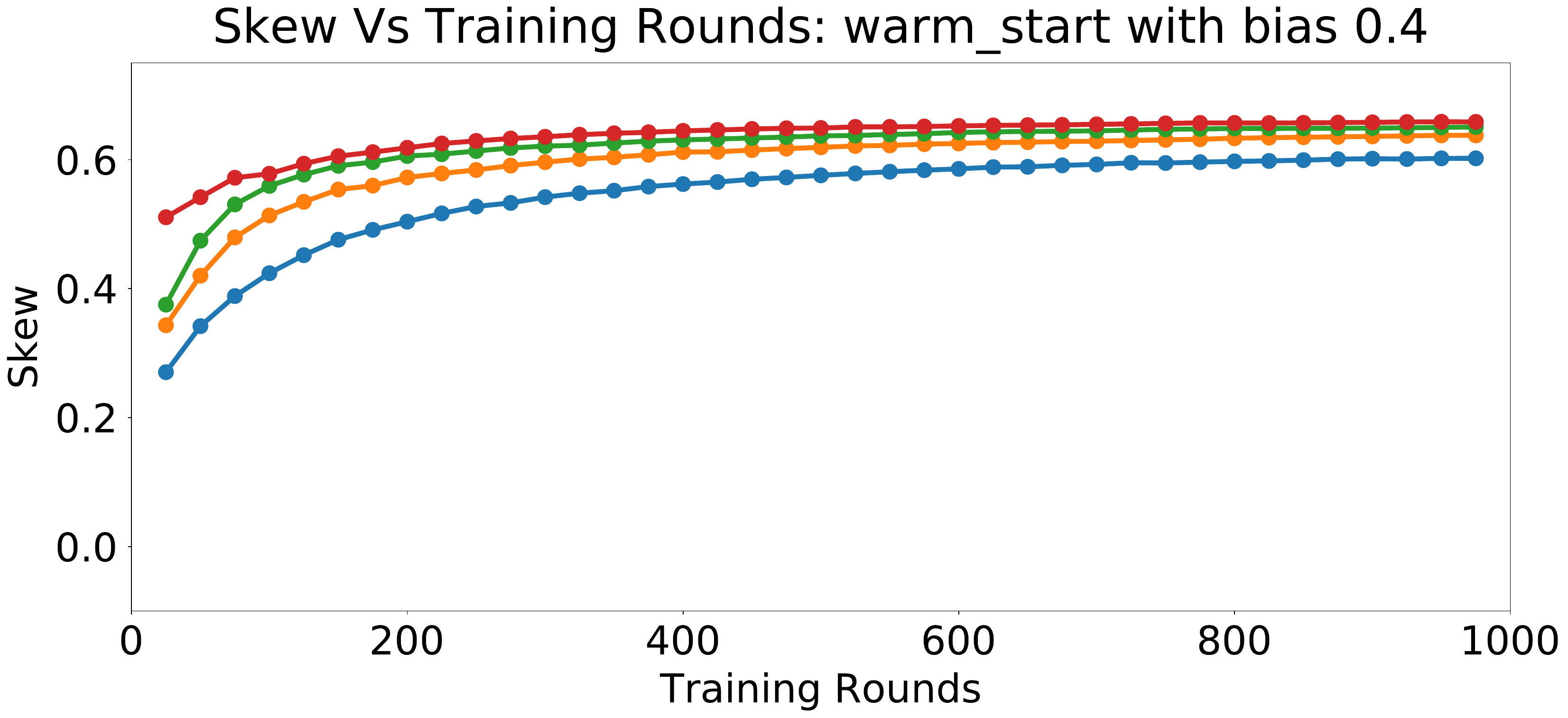}
        \caption{Evolution with fixed bias}
        \label{fig:evolution1}
    \end{subfigure}
    
    \hfill
    
    \begin{subfigure}[b]{0.85in}
        \includegraphics[width=0.95in]{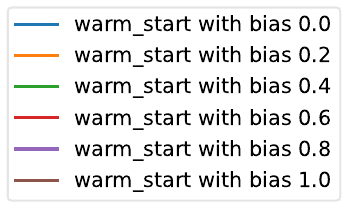}
        \caption{Legend (fixed learning rate)}
        \label{fig:evolution4}
    \end{subfigure}
    \quad \begin{subfigure}[b]{2.3in}
        \includegraphics[width=2.3in]{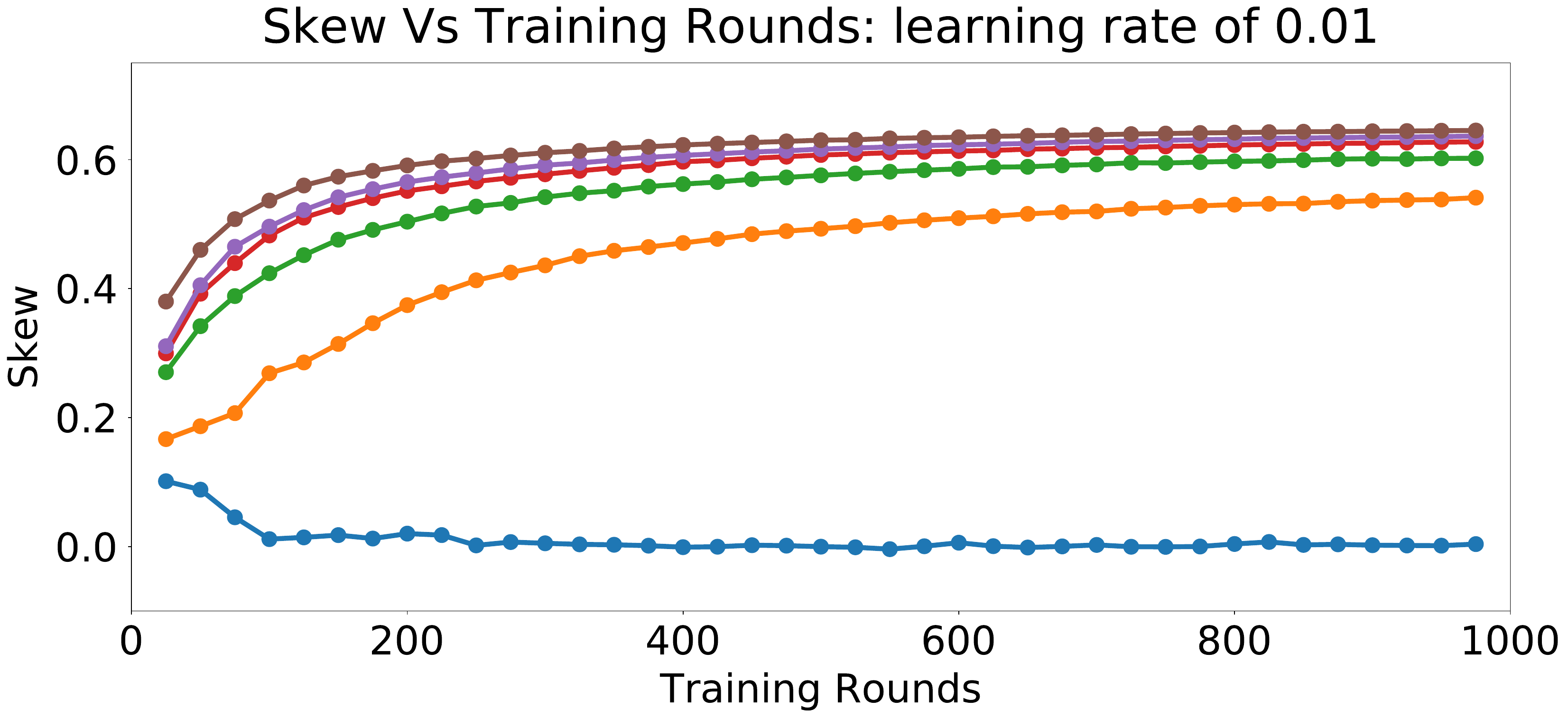}
        \caption{Evolution with fixed learning rate}
        \label{fig:evolution3}
    \end{subfigure}
\caption{Evolution of Skew in Online Personalization}\label{fig:evolution}
\end{figure}

In Figure \ref{fig:evolution1}, we evaluate the effect of learning rate and the number of recommendations on Skew$_{group = 1}$@25 metric for personalizing towards the preference of a single user with $p_{bias} = 0.4$. Since the model learns the biased preferences of the user more and more with each subsequent training round, we observe that the skew increases with the number of training rounds. Furthermore, as the learning rate increases, the skew increases (keeping the number of rounds fixed), since we are moving the model ``faster'' towards the biased preferences of the user.

In Figure \ref{fig:evolution3}, we look into the effect of $p_{bias}$ and the number of recommendations on Skew$_{group = 1}$@25 metric with a fixed learning rate of $0.01$. We observe that the skew increases with the number of rounds (recommendations), as we learn the biased behavior of the user. Also, as $p_{bias}$ increases, the skew grows faster with more number of rounds. For the model with $p_{bias} = 0$, the weights do not change much with more training rounds. Hence, the skew does not change significantly with the increase in the number of training rounds. However, for users with high $p_{bias}$, the feature weights incorporate the algorithmic bias more and more as the number of training rounds increases and hence the skew increases. Overall, this study shows that for biased users, it is really easy to erase the unbiased/neutral nature of the warm-start model, if the goal is to understand and personalize the model to their (biased) preferences.

\section{Fair Regularization} \label{sec:fair_regularization}
We next discuss our method for restricting the learned models to be fair, followed by an evaluation of this approach in \S\ref{sec:experiment2}.

Consider the problem of constructing a fair machine learning model. Assume that we have data points in the form of (a feature column vector, scalar protected attribute, scalar label) tuple, $(x, a, y)$. Let us denote our training data as S =$\{$ $(x_1, a_1, y_1)$, $ (x_2, a_2, y_2)$, $ (x_3, a_3, y_3)$, $\ldots$, $ (x_N, a_N, y_N)$ $\}$.

We would like to learn a model $h(x) = w^Tx$ such that $h(x)$ and $a$ are mutually independent. Let us define matrix $X$, $Y$, $A$ out of training data, formed by stacking $x_i$s, $y_i$s, $z_i$s as columns.
$$
X = 
\begin{bmatrix}
x_1 & x_2 & x_3 & \ldots & x_N
\end{bmatrix} 
$$
$$
A = 
\begin{bmatrix}
a_1 & a_2 & a_3 & \ldots & a_N
\end{bmatrix}
$$
$$
Y = 
\begin{bmatrix}
y_1 & y_2 & y_3 & \ldots & y_N
\end{bmatrix}
$$
We assume that $S$ is given to us and it can be used to select the $h \in H$ as per our desired constraints. Such a choice of $h$ only depends on $S = \{ (x_i, a_i, y_i) | i = 1,2,3,\ldots,N \}$ and that data points in the form of $(x,a,y)$ can be sampled, independent and identically distributed from $D$  and are especially independent of $S$. Such data points $(x, a, y)$ would also be independent of the selected $h$ which solely depends on $S$.

Now we present our fair learning problem for linear models as:
$$
\textbf{Minimize}_w \text{  } ||Y - w^TX||^2  
$$
\vspace{-5mm}
$$
\textbf{s.t. } \text{the distributions of } w^Tx \perp a 
$$
To make this problem more tractable, we make two relaxations.
\paragraph{\textbf{Independence to Correlation:}}We first relax the constraint of $w^Tx$ and $a$ being independent to only being uncorrelated. It is well-known that independent variables are uncorrelated. Hence, zero correlation is a weaker constraint than independence. In the special case of jointly Gaussian variables, being uncorrelated implies independence. Further, we use empirical correlations computed from training data, in the place of actual correlations in the optimization. 

\paragraph{\textbf{a to Model Estimate of a:}} Instead of using $a$ in the optimization, we first learn a estimate of $\hat a$ from $x$ using a similar model used to learn $y$ from $x$ and then use those estimates in the optimization. Such estimate $\hat a$ is the best that the model can possibly infer from the training data about the protected attribute. We try to find the best fit least square solution to the equation $w_a^TX = A$. Once we learn the best fit (in least squares sense) linear model to predict $A$ from $X$, we can approximate the correlation with $\frac{1}{N} w_a^TX (w^TX)^T = \frac{1}{N} w_a^TXX^T w = w_a^T \text{Cov}(X, X) w = w_a^T \Sigma_{x} w$.\\

With the above relaxations, learning a fair linear model reduces to solving:
$$
\textbf{Minimize}_{w_a} \text{  } ||A - w_a^TX||^2  
$$
\vspace{-5mm}
$$
\textbf{ s.t. }
||w_a|| < \epsilon_a
$$ 
and,
$$
\textbf{Minimize}_w \text{  } ||Y - w^TX||^2  
$$
\vspace{-5mm}
$$
\textbf{ s.t. }
||w^T \Sigma_x w_a|| < \epsilon
$$ 
where $\epsilon_a$ and $\epsilon$ are sufficiently small real numbers.

The above optimization problem has two parts. We first learn a best fit linear model from $X$ to $A$. We use the weights learned in the first model to optimize for a fair best fit linear model from $X$ to $Y$.
$$\textbf{Minimize  }||A - w_a^TX||^2$$ 
$$\textbf{ s.t. } ||w_a|| < \epsilon_a$$ 
Note that the first minimization problem has an additional constraint of small L2 norm on weights $w_a$ because the unconstrained optimization might be ill-posed. The contribution of the harmless attribute components in feature vectors' $x_i$s towards the protected attribute is the ill-posed part. Hence, to make the contributions of the harmless attribute components in feature vectors $x_i$s small, we impose the additional constraint of small L2 norm on the weights $w_a$.

The second optimization (for fair linear model) can be equivalently expressed as: 
$$\textbf{Minimize  }||Y - w^TX||^2 + \lambda ||w^T w_{reg}||^2 ~,$$ 
where $w_{reg} = \Sigma_x w_a$. This is the Langrange Multiplier version or the regularized linear regression. We call this kind of regularization as Fair Regularization from now on. This can be solved exactly as:
$$(XX^T + \lambda w_{reg}w_{reg}^T)w = XY^T ~,$$
or it can also be solved iteratively by a gradient update equation:
$$w_{n+1} = w_{n} + \eta(y_{n} - \text{sgn}(w_{n}^Tx_{n}))x_n - \lambda(w_n^Tw_{reg})w_{reg} ~.$$

Posing the fair learning problem as a regularized model has the benefit that:
\begin{itemize}
	\item We need not transform the data to any kind of fair space. Fairness conditions are imposed on the model at train time, and not on training data which can still be used in its raw potentially unfair form.
	\item Fairness is imposed in an end-to-end single module compared to transforming the data to a fair space and then using the transformed data for the learning task at hand.
\end{itemize}

\subsection{Extensions}
\textbf{Extending to other fairness conditions:}
Some commonly used fairness notions in machine learning are described in Appendix \ref{sec:fairnessnotions}. To achieve other kinds of fairness conditions like conditional independence $( C \perp A \mid Y)$, our method can be tweaked by replacing the expectations with conditional expectations in the constraint of the second optimization problem. The resulting equation would result in $\Sigma_{x|y} = \mathbb{E}[XX^T \mid Y]$ in the place of $\Sigma_x = \mathbb{E}[XX^T]$. However, within the scope of our paper, we only work with the independence constraint. 

~

\noindent \textbf{Extending to more general hypothesis space:}
Finally, we would like to describe how our approach can be extended for more general hypothesis space. For general hypothesis space $H$, our first learning problem can be extended to learning the best hypothesis $h_a \in H$ which can predict $a_i$ from $x_i$ for $i = 1, 2, 3, \ldots, N$. Then in the second learning problem, we extend our regularization term which gets optimized with the training loss using the following insight:
$$w^Tw_{reg} = w^T \Sigma_x w_a = \mathbb{E}[(w^Tx)(w_a^Tx)^T] = \mathbb{E}[(h(x))(h_a(x))^T]$$
Hence, while training to reduce the loss $L(h_n(x_n),y_n)$, we replace the loss with $L(h_n(x_n), y_n)$ + $\lambda$ $(h_n(x_n))(h_a(x_n)^T)^2$
where $h_n$ is the hypothesis used and $x_n$ is the training data point chosen in the $n^{th}$ training iteration. Such a regularization has the effect that it penalizes the correlation of the model with another model that is the best fit that can be learned to predict the protected attribute. However, such a regularization is more noisy in general and takes more training rounds to converge compared to our case of linear model where the correlation can be better calculated and used at every iteration. Within the context of this paper, we only work with linear models, and leave such extensions to future work.

\begin{figure*}[!t]
    \centering
    \begin{subfigure}[b]{1.6in}
        \includegraphics[width=1.6in]{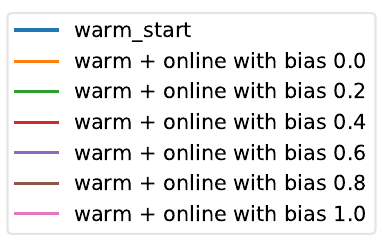}
        \caption{Legend}
        \label{fig:reg_skew_legend}
    \end{subfigure}
    \quad \begin{subfigure}[b]{2.2in}
        \includegraphics[width=2.2in]{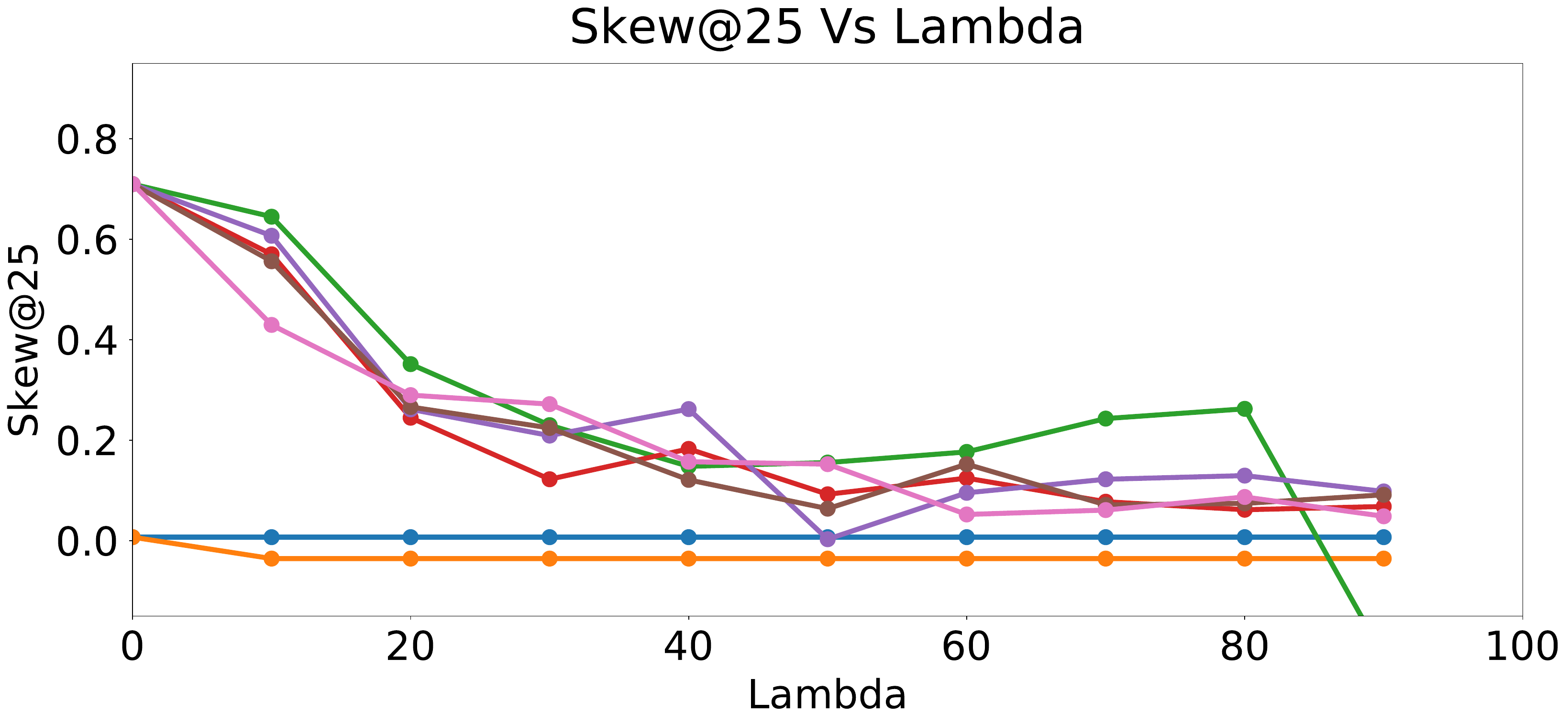}
        \caption{Regularized Skew$_{group = 1}$@25}
        \label{fig:reg_skew25}
    \end{subfigure}
    \quad \begin{subfigure}[b]{2.2in}
        \includegraphics[width=2.2in]{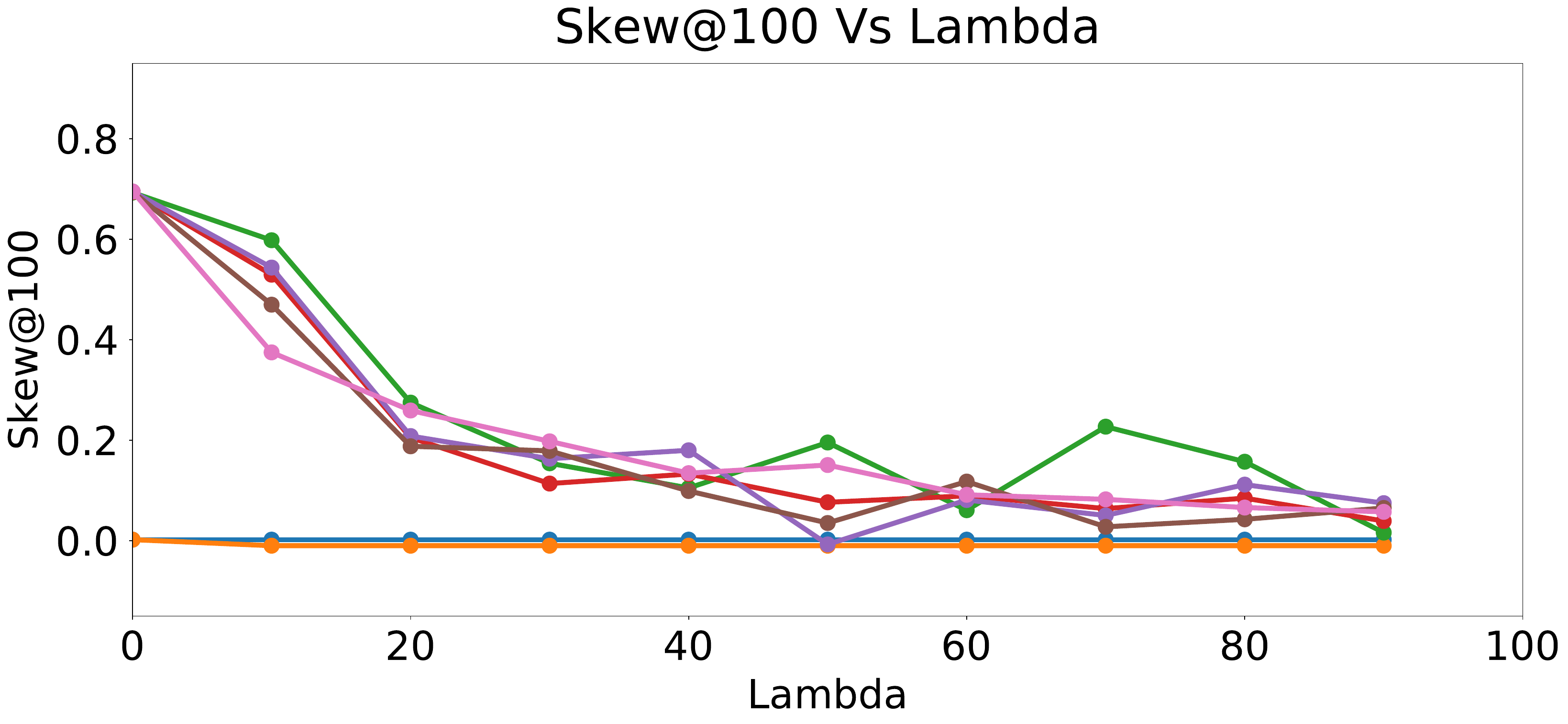}
        \caption{Regularized Skew$_{group = 1}$@100}
        \label{fig:reg_skew100}
    \end{subfigure}
    
    \hfill
    
    \begin{subfigure}[b]{2.2in}
        \includegraphics[width=2.2in]{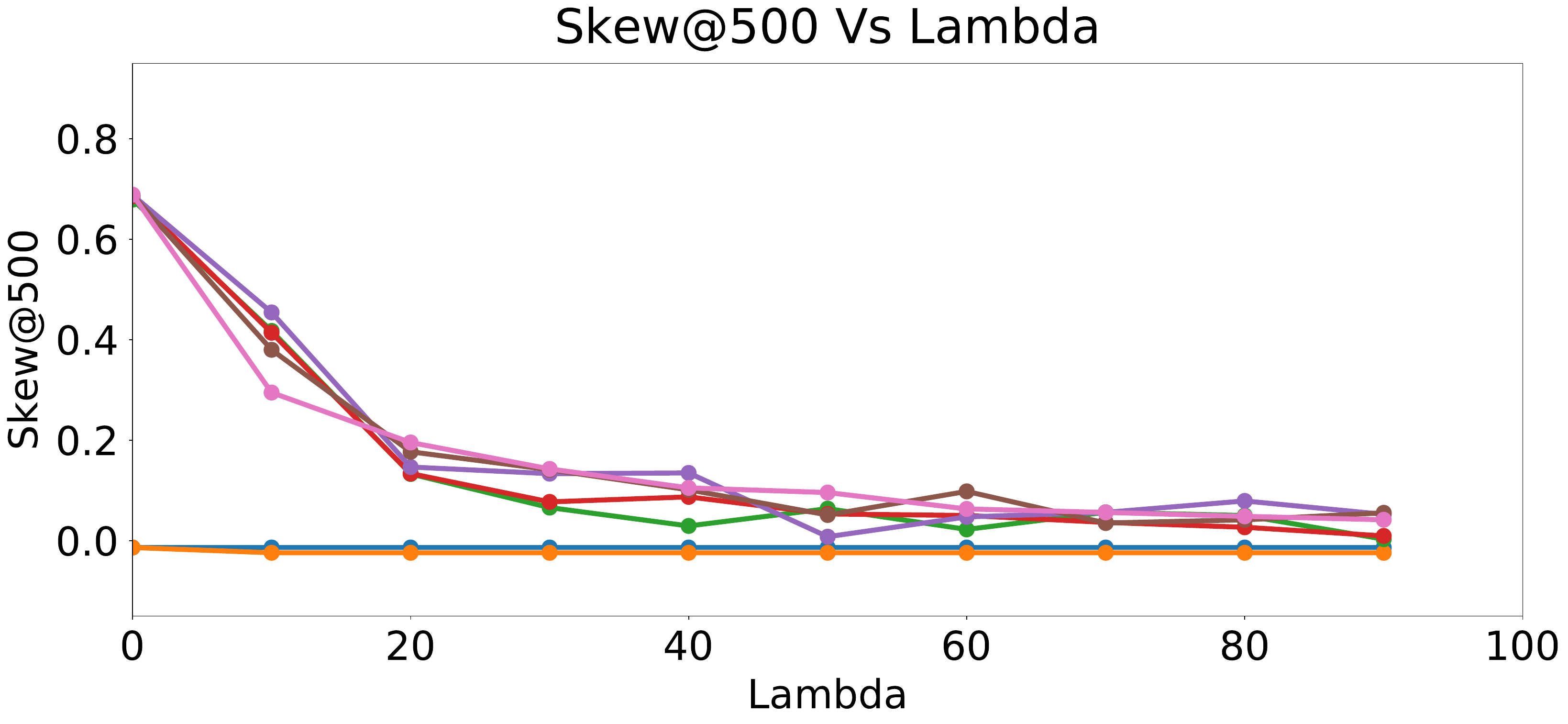}
        \caption{Regularized Skew$_{group = 1}$@500}
        \label{fig:reg_skew500}
    \end{subfigure}
    \quad \begin{subfigure}[b]{2.2in}
        \includegraphics[width=2.2in]{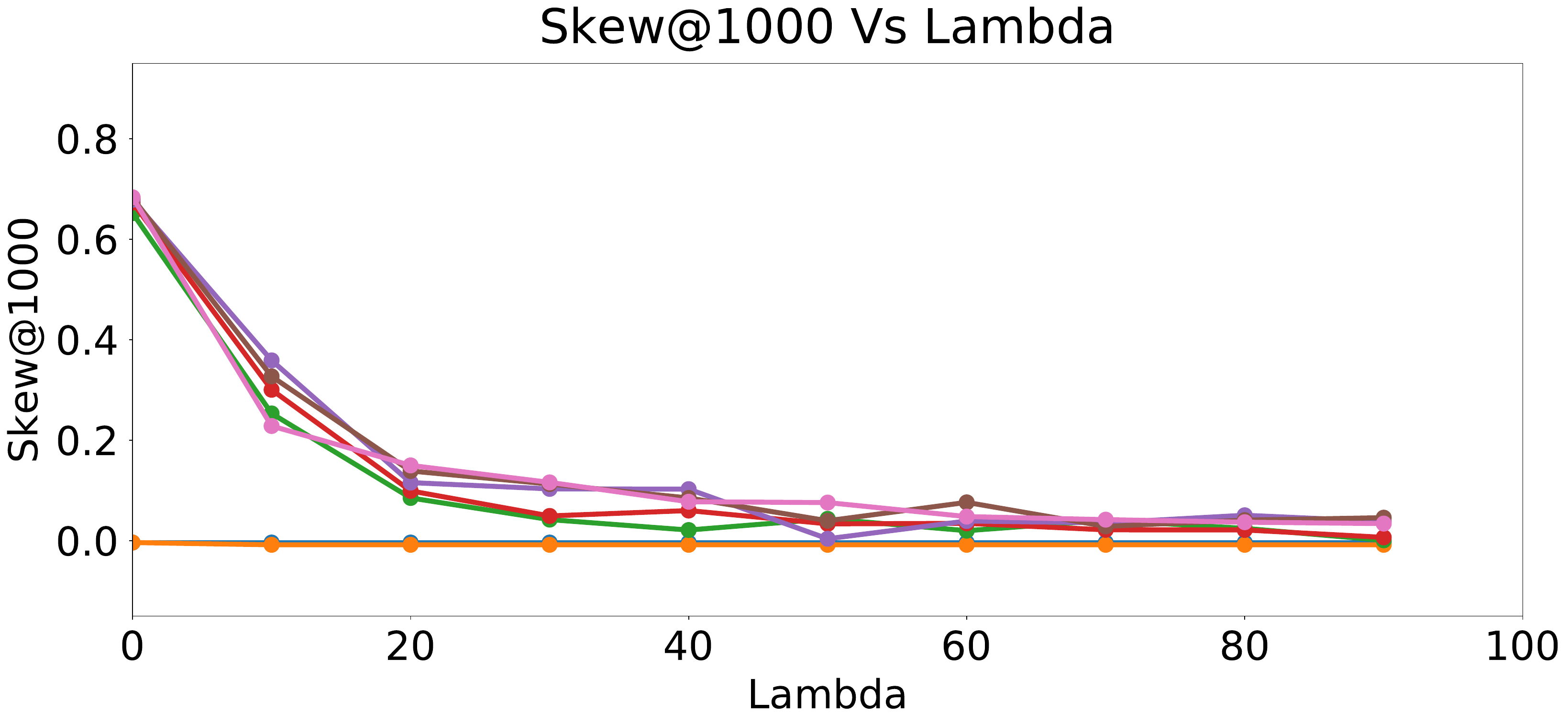}
        \caption{Regularized Skew$_{group = 1}$@1000}
        \label{fig:reg_skew1000}
    \end{subfigure}
    \quad \begin{subfigure}[b]{2.2in}
        \includegraphics[width=2.2in]{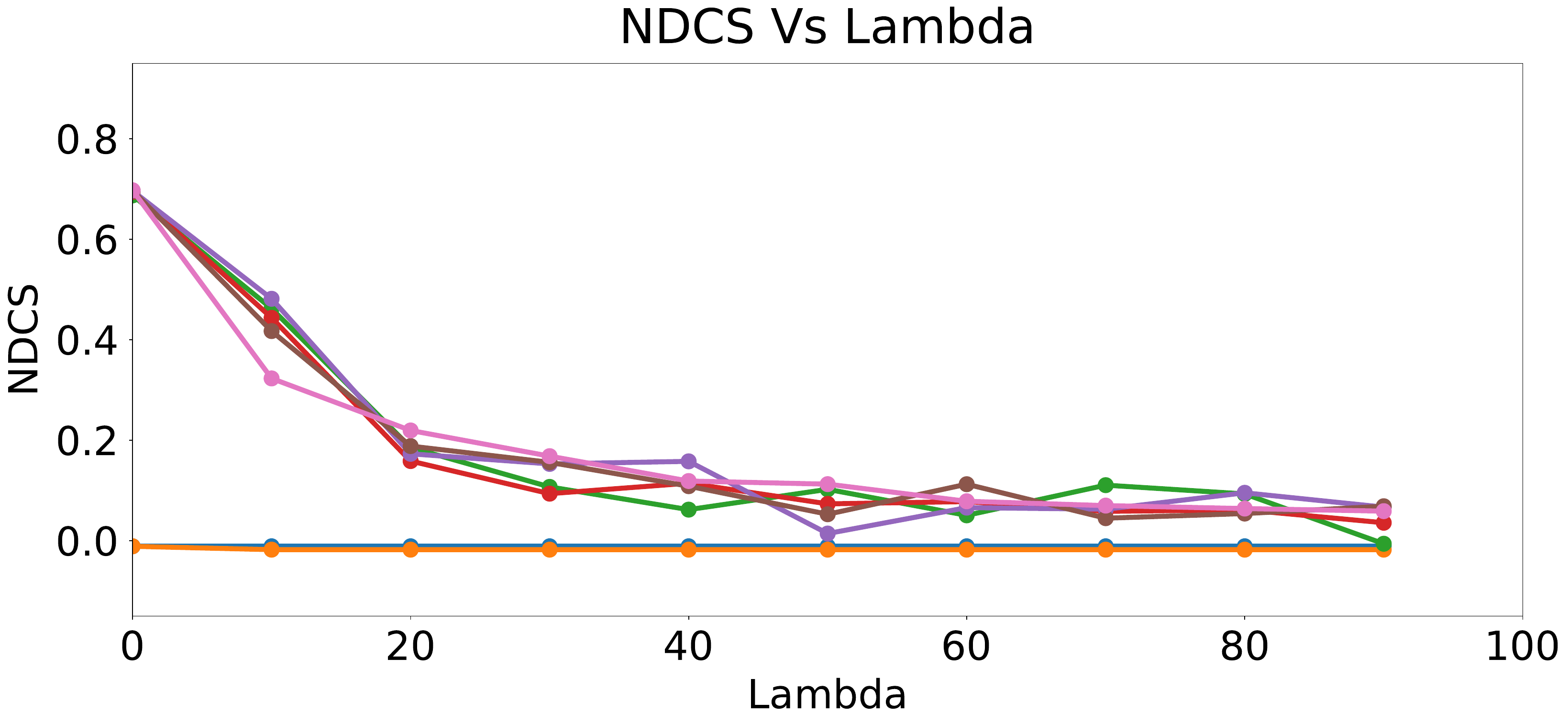}
        \caption{Regularized NDCS}
        \label{fig:reg_ndcs}
    \end{subfigure}
    \caption{Regularized Skew and NDCS}\label{fig:reg_skew}
\end{figure*}

\begin{figure*}[!t]
    \centering
    \begin{subfigure}[b]{1.6in}
        \includegraphics[width=1.6in]{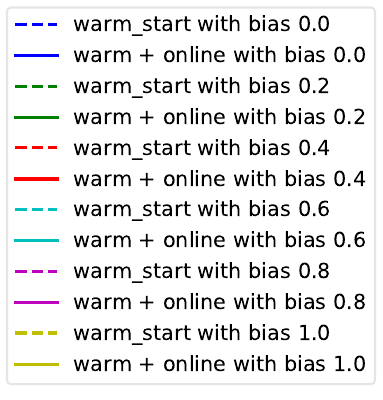}
        \caption{Legend}
        \label{fig:reg_precision_legend}
    \end{subfigure}
    \quad \quad \begin{subfigure}[b]{5.0in}
        \begin{subfigure}[b]{2.2in}
            \includegraphics[width=2.2in]{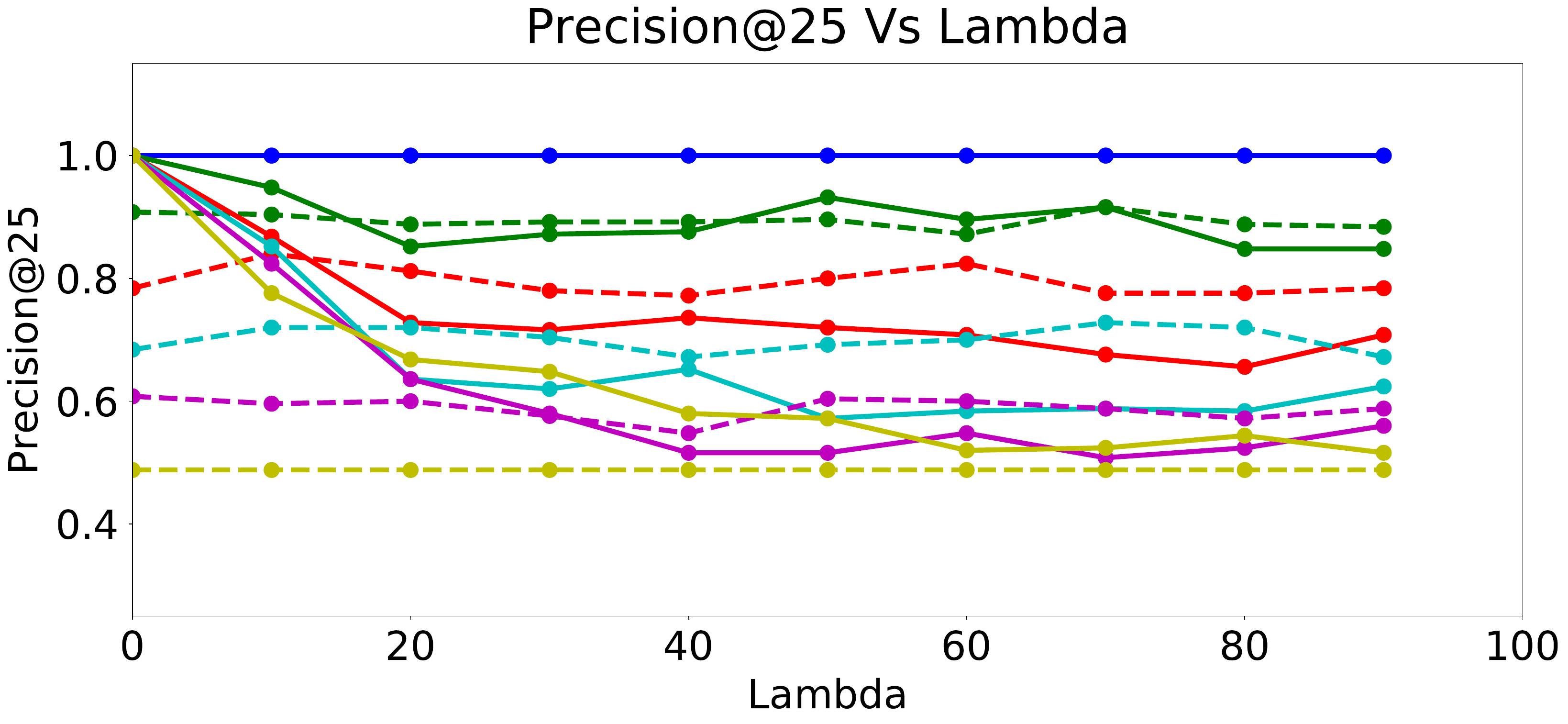}
            \caption{Regularized Precision\cut{$_{group = 1}$}@25}
            \label{fig:reg_precision25}
        \end{subfigure}
        \quad \begin{subfigure}[b]{2.2in}
            \includegraphics[width=2.2in]{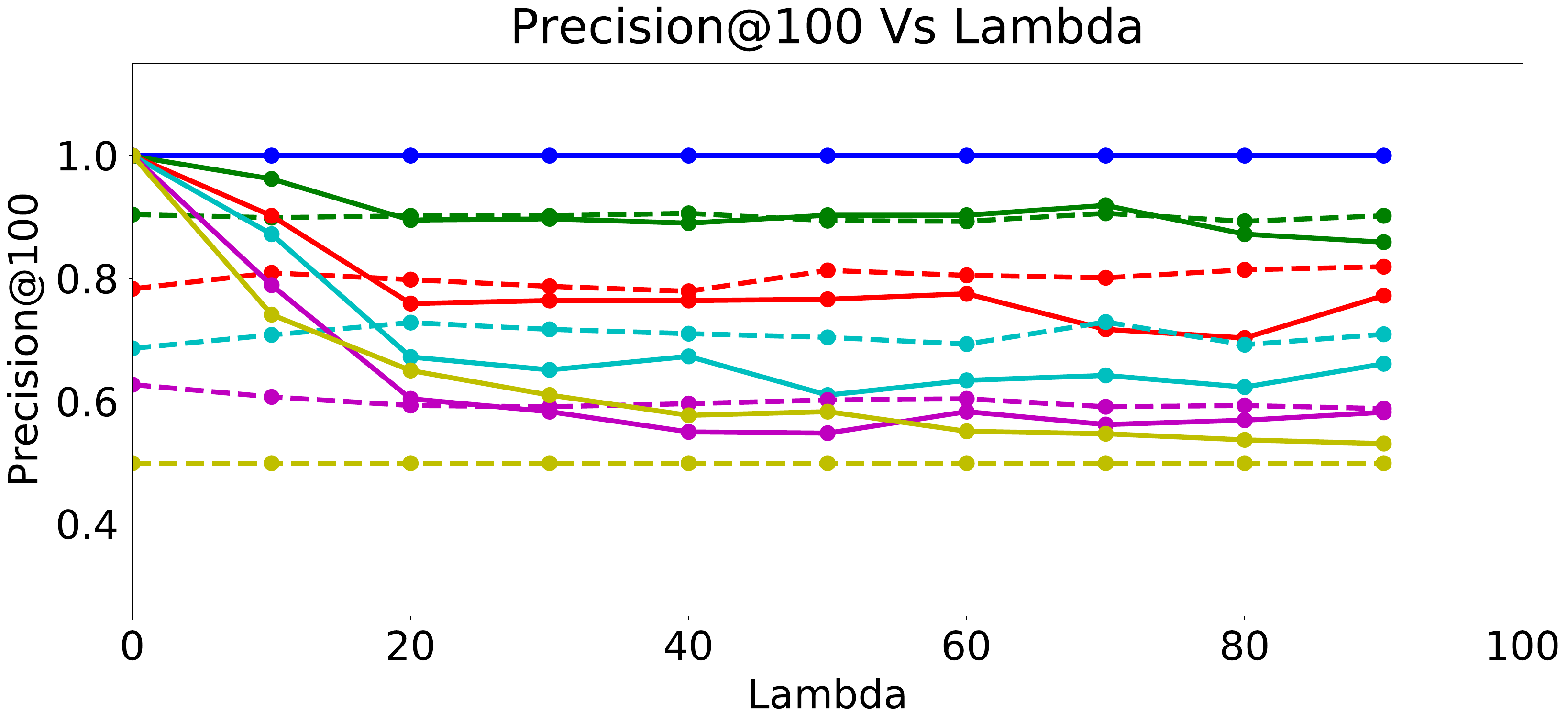}
            \caption{Regularized Precision\cut{$_{group = 1}$}@100}
            \label{fig:reg_precision100}
        \end{subfigure}
        
        \hfill
        
        \begin{subfigure}[b]{2.2in}
            \includegraphics[width=2.2in]{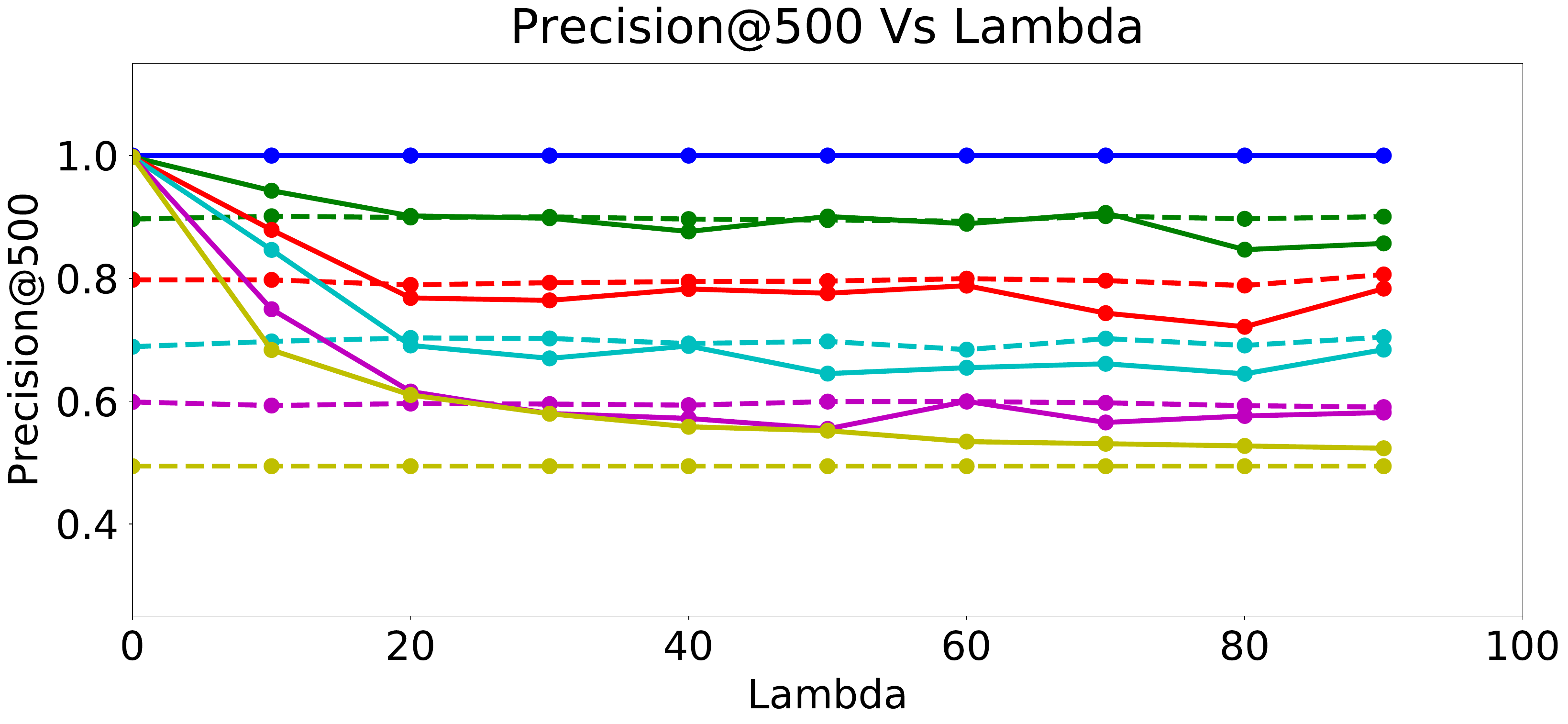}
            \caption{Regularized Precision\cut{$_{group = 1}$}@500}
            \label{fig:reg_precision500}
        \end{subfigure}
        \quad \begin{subfigure}[b]{2.2in}
            \includegraphics[width=2.2in]{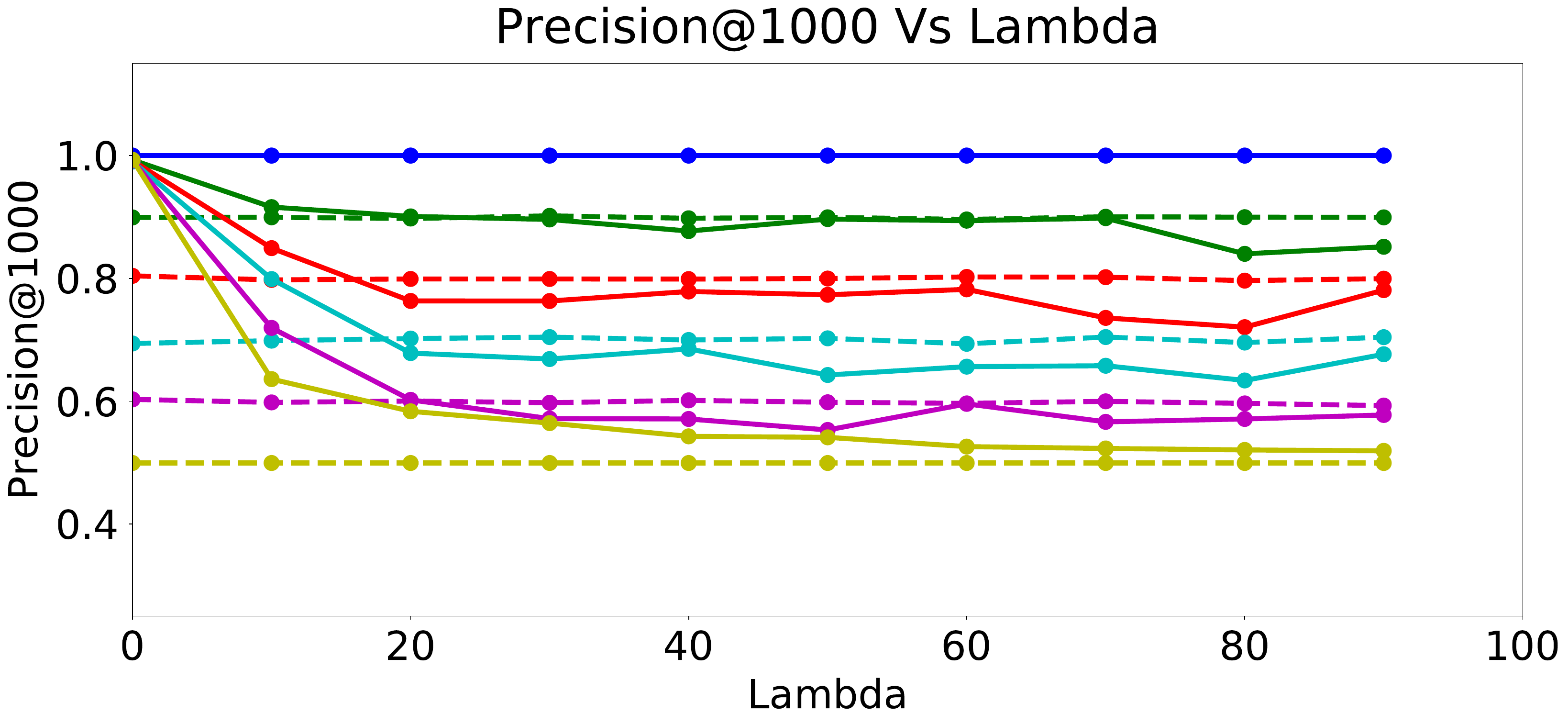}
            \caption{Regularized Precision\cut{$_{group = 1}$}@1000}
            \label{fig:reg_precision1000}
        \end{subfigure}
    \end{subfigure}
    \caption{Regularized Precision}\label{fig:reg_precision}
\end{figure*}

\subsection{Evaluation of Mitigation using Fair Regularization} \label{sec:experiment2}
We next build upon our experiments in \S\ref{sec:fairness} with the fair regularization idea developed in \S\ref{sec:fair_regularization}. We study the effect of the regularization parameter $\lambda$ on skew and precision. We follow the same setup as in \S\ref{sec:experiment1} and \S\ref{sec:fairness} except that we fix a constant learning rate  and experiment with different values of $\lambda$.

Figure \ref{fig:reg_skew} presents Skew@k and NDCS results as function of $\lambda$. As expected, Skew@k and NDCS drop towards $0$ for all data sets generated with different amount of user bias as $\lambda$ increases, which shows the efficacy of the fair regularization approach. However, we incur a trade-off in precision of the recommendations, since the regularization reduces the ability of the models to learn the biased preferences of the users. From Figure \ref{fig:reg_precision}, we can see that as $\lambda$ increases, Precision@k of the online updated model drops to Precision@k of the warm-perceptron model for each type of user with different $p_{bias}$. These results suggest an inherent tension between ensuring fairness and personalization. We could think of fair regularization also as attempting to address the over-fitting problem. Here, our goal is to prevent or minimize overfitting to the biased behavior of the user by imposing constraints on the learned weights for the proxy attributes. However, since the main objective of personalization is to learn and cater to the preferences of the given user as closely as possible (hence, for lack of a better word, ``overfit'' to the user's preferences, however biased they may be), regularization for ensuring fairness could hurt precision as we observed. The lesson learned is that we need to be careful when designing online personalized models and be aware of their potential to introduce algorithmic bias, even more prominently compared to offline models which may have been trained on data from a large set of users and tested for desired fairness properties.

%\vspace{-0.2in}
\section{Conclusion}\label{sec:conclusion}

Considering the importance of understanding and addressing algorithmic bias in personalized ranking mechanisms, we formulated the problem of studying fairness in online personalization. We performed an investigation of potential biases possible in online personalization settings involving ranking of individuals.
Starting from a fair warm-start machine learned model, we empirically showed that online personalization can cause the model to learn to act in an unfair manner, if the user is biased in his/her responses. As a part of this study, we presented a stylized mathematical model for generating training data with potentially biased features as well as potentially biased labels, and quantified the extent of bias that is learned by the model when the user sometimes responds in a biased manner as in plausible real-world settings. The ideas/intuition underlying our framework for generating training data for simulation purposes are likely to be of broader interest to the research community for evaluating fairness-aware algorithms in other scenarios, where either suitable datasets may not be available or it may be infeasible to perform extensive experimentation with real datasets. We then formulated the problem of learning personalized models under fairness constraints, and proposed a regularization based approach for mitigating biases in machine learning, specifically for linear models. We evaluated our methodology  through extensive simulations with different parameter settings.

\begin{acks}
The authors would like to thank other members of LinkedIn Careers and Talent Solutions teams for their collaboration for performing our experiments, and 
Deepak Agarwal,
Erik Buchanan,
Patrick Cheung,
Patrick Driscoll,
Nadia Fawaz, 
Joshua Hartman,
Rachel Kumar,
Ram Swaminathan,
Hinkmond Wong, 
Ryan Wu,
Lin Yang,
Liang Zhang,
and
Yani Zhang
for insightful feedback and discussions.
\end{acks}

{
\bibliographystyle{abbrv}
\bibliography{paper}
}

\appendix
\section{Appendix\cut{Background}} \label{sec:background}
\subsection{Legal Frameworks on Bias and Discrimination}\label{sec:legal}
Over the last several decades, legal frameworks around the world have prohibited the use of sensitive attributes such as race, gender, sexual orientation, and age for different forms of decision making. We refer to such attributes whose use in certain decision making has been prohibited by law as \emph{protected attributes}. Examples of attributes declared as protected by law in the United States include: 
\begin{itemize}
	\item Race, Color, Sex, Religion, National origin (Civil Rights Act of 1964; Equal Pay Act of 1963),
	\item Age (Age Discrimination in Employment Act of 1967),
	\item Disability status (Rehabilitation Act of 1973; Americans with Disabilities Act of 1990), and
	\item Veteran status (Vietnam Era Veterans' Readjustment Assistance Act of 1974).
\end{itemize}

When legal frameworks prohibit the use of such protected attributes in decision making, there are usually two competing approaches on how this is enforced in practice: \emph{Disparate Treatment} vs. \emph{Disparate Impact}. Avoiding disparate treatment requires that such attributes should not be actively used as a criterion for decision making and no group of people should be discriminated against because of their membership in some protected group. Avoiding disparate impact requires that the end result of any decision making should result in equal opportunities for members of all protected groups irrespective of how the decision is made. This can often mean that certain forms of affirmative action based on protected attributes may be needed to encourage or improve opportunities for certain protected groups to offset the biases present in the society due to historical reasons. Our work can be thought of as following the approach of avoiding disparate impact.

\subsection{Sources of Bias}\label{sec:biassources}

To prevent a machine learning model from incorporating biases, it is not enough to prohibit the usage of protected attributes. Often one or more protected attributes could be correlated with seemingly harmless attributes (hereafter denoted as ``proxy attributes''). 
Presence of proxy attributes in a machine learning model potentially allows it to extract information on protected attributes and further use it to make decisions in an unfair manner. 

Some possible reasons for biases in the outcomes of machine learning models include:
\begin{itemize}
\item Presence of proxy attributes in the set of features utilized for training,
\item Different quality and quantity of samples from different protected groups in training data, and
\item Structured noisy labels in training data with different amounts of noise depending upon protected attributes.
\end{itemize}
In \S\ref{sec:data}, we describe our stylized model of simulating training data which embodies most of the above sources of algorithmic bias.

\subsection{Different Notions of Fairness} \label{sec:fairnessnotions}

\begin{figure}[!t]
	\centering
	\includegraphics[width=3.3in]{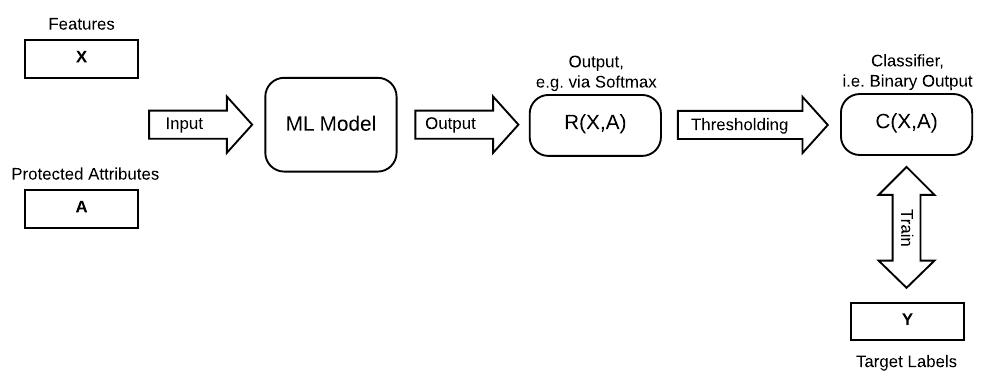}
	\caption{Formal Machine Learning Model}
	\label{fig:ml-model}
\end{figure}

Consider the formal setup of a machine learning model shown in Figure \ref{fig:ml-model}. Let $X$ be the feature vector that is used by the machine learning model as input and $A$ be the corresponding protected attribute, which, for simplicity, is assumed to be a binary variable taking values 1 and 0 (e.g., whether sex $=$ `female'; whether age $\ge 40$). Let $R(X, A)$ be the output of the model, which can be thought of as the result of a softmax procedure, and in the range $(0,1)$. A thresholding scheme is applied on the output $R(X, A)$ to convert it into binary labels $C(X, A)$. The model is trained such that $C(X, A)$ follows the binary labels $Y$ in training data. In such a setting, we could consider multiple approaches to formalize what we mean by fairness, leading to the following definitions~\cite{nips_2017_tutorial}.
\begin{enumerate}
	\item $C \perp A$ :  Independence condition requires that output is independent of the protected attribute. This is often called statistical parity where chances of ``Accept'' (or ``Reject'') conditioned on the value of the protected attribute are the same for both protected groups, i.e.,
	$$P(C = 1 \mid A = 1) = P(C = 1 \mid A = 0) = P(C = 1).$$
	Many approximate versions of this are often recommended by guidelines such as the Uniform Guidelines for Employee Selection Procedures, adopted by certain U.S. government agencies\cut{ (EEOC, Department Of Labor, Department of Justice, and the Civil Service Commission)} in 1978. According to these guidelines, adverse or disparate impact is determined by using the ``four-fifths or 80\% rule'' which requires that the selection rate for any race, sex, or ethnic group be at least four-fifths (or 80\%) of the rate for the group with the highest rate for the process to be considered as not having adverse impact. For the case of binary protected attribute, this corresponds to 
	$$\frac{1}{0.8} \geq \frac{P(C = 1 \mid A = 1)}{P(C = 1 \mid A = 0)} \geq 0.8.$$
	A potential problem with this criterion is that often training data might not support equal ``Accept'' chances for all protected groups and hence enforcing such an independence constraint might hurt the precision or other performance measures of the algorithm.
	\item $C \perp A \mid Y$ : Conditional independence condition requires that the model makes equal amounts of mis-classifications for all protected groups, i.e., 
	$$P_{A=1}(C = 1 \mid Y = 1) = P_{A=0}(C = 1 \mid Y = 1),$$
	$$P_{A=1}(C = 1 \mid Y = 0) = P_{A=0}(C = 1 \mid Y = 0).$$
	Thus, it translates to enforcing equal true positive rates and false positive rates for all protected groups. This criterion can allow the model training to proceed to fit to the data\cut{ perfectly} with high precision.
	\item $Y \perp A \mid C$ : This condition is similar to the previous condition, but with the roles of $C$ and $Y$ reversed. Note that $C$ is trained to follow $Y$ in training data and hence it makes sense to treat them equivalently. 
\end{enumerate}  
In this paper, we focus on the independence condition, i.e., (1) above, based on which we devised an approach to mitigate bias in \S\ref{sec:fair_regularization}.

\end{document}